\documentclass[runningheads]{llncs}
\usepackage[T1]{fontenc}
\usepackage{graphicx}
\usepackage{color}
\usepackage{needspace}

\makeatletter
\renewcommand\subsubsection{\@startsection{subsubsection}{3}{\z@}%
  {-10\p@ \@plus -4\p@ \@minus -4\p@}%
  {0.5em \@plus 0.22em \@minus 0.1em}%
  {\normalfont\normalsize\bfseries}}
\makeatother

\newcommand{\mytitle}{Tractability Frontiers in Multi-Robot Coordination and Geometric Reconfiguration\thanks{Work by T.G. and D.H. has been supported in part by the Israel Science
Foundation (grant no~2261/23),
by NSF/US-Israel-BSF (grant no.~2019754),
by the Blavatnik Computer Science Research Fund, by the Israeli Smart Transportation Research Center,
and by the Shlomo Shmelzer Institute for
Smart Transportation at Tel Aviv University.}}
\title{\mytitle} 

\titlerunning{Tractability Frontiers in Multi-Robot Coordination} %

\author{Tzvika Geft\inst{1}\thanks{Work by T.G. was carried out while the author was at Tel Aviv University.}
\and
Dan Halperin\inst{2}
\and
Yonatan Nakar\inst{2}}
\institute{Rutgers University, New Brunswick NJ, USA  \and Tel Aviv University, Israel}

\authorrunning{T. Geft et al.}

\usepackage [autostyle, english = american]{csquotes}
\MakeOuterQuote{"}

\usepackage[sort,nocompress]{cite} %

\usepackage{multirow}
\usepackage{makecell}

\usepackage{amsmath,amsthm}
\usepackage{thm-restate} %
\usepackage{hyperref} %
\usepackage{cleveref} %
\usepackage{subcaption}
\usepackage{color}
\usepackage{comment}
\usepackage{mathtools}
\usepackage{MnSymbol}
\usepackage{float}
\usepackage{amsfonts}
\usepackage[linesnumbered,ruled,vlined,algo2e]{algorithm2e}
\usepackage{enumitem}
\usepackage{tikz}
\usetikzlibrary{positioning, calc}
\usetikzlibrary{arrows.meta}

\usepackage{environ}
\newif\ifappendix
\NewEnviron{maybeappendix}[1]
    {\ifappendix
        \expandafter\global\expandafter\let\csname putmaybeappendix#1\endcsname\BODY%
    \else
        \expandafter\newcommand\csname putmaybeappendix#1\endcsname{}\BODY%
    \fi}
\newcommand{\putmaybeappendix}[1]{\csname putmaybeappendix#1\endcsname}

\usepackage{ifthen}
\newboolean{ispaper} 
\setboolean{ispaper}{false} %

\newboolean{isfinalpaper} 
\setboolean{isfinalpaper}{true} %

\newcommand{\src}[1] {\ensuremath{s(#1)}}
\newcommand{\trg}[1] {\ensuremath{t(#1)}}
\newcommand{\pth}[1]{ %
  \ifx#1\relax \ensuremath{\pi} \else \ensuremath{\pi_{#1}} \fi
}

\newcommand{\bef}{\ensuremath{B}\xspace}
\newcommand{\aft}{\ensuremath{A}\xspace}
\newcommand{\ppath}{\ensuremath{P_{\bef,\aft}}}

\newcommand{\set}[1]{\ensuremath{\{#1\}}}

\definecolor{gray}{rgb}{0.35,0.35,0.35}
\definecolor{blue}{rgb}{0,0,1}
\definecolor{red}{rgb}{1,0,0}
\definecolor{orange}{rgb}{0.75, 0.4, 0}
\definecolor{green}{rgb}{0.0, 0.5, 0.0}

\newcommand{\tzvika}[1]{{\color{blue}\textbf{Tzvika: }\sf#1}}

\newcommand{\danny}[1]{{\color{red}\textbf{Danny: }\sf#1}}

\def\P{\mathcal{P}} \def\C{\mathcal{C}} 
  
 \def\I{\mathcal{I}} 
  \def\I{\mathcal{I}}
\def\T{\mathcal{T}}  
  \def\D{\mathcal{D}}
 \def\R{\mathcal{R}} 
\def\X{\mathcal{X}} \def\A{\mathcal{A}} \def\Y{\mathcal{Y}}

\def\E{\mathcal{E}}
\def\Z{\mathcal{Z}}

\def\dR{\mathbb{R}}

\renewcommand{\tzvika}[1]{}
\renewcommand{\danny}[1]{}

\usepackage{microtype}

\newcommand{\MSRsq}{MSR$(\square)$\xspace}
\newcommand{\MSRnear}{MSR$(\square, \scalebox{0.83}{$\square$})$\xspace}

\newcommand{\msrdisks}{\ensuremath{M}\xspace}

\newcommand{\motiong}{\ensuremath{G}\xspace}
\newcommand{\motiongdisks}{\ensuremath{\hat{G}}\xspace}
\newcommand{\lcsdisks}{\ensuremath{H}\xspace}
\newcommand{\solvable}{spannable by a forest of full binary trees}
\newcommand{\shortsolvable}{SFFBT\xspace}

\newcommand{\thinpolys}{is composed of thin polyominoes\xspace}

\newcommand{\noObstBounded}[2]{
    \begin{figure}[#1]
        \centering
        \includegraphics[width=#2\textwidth]{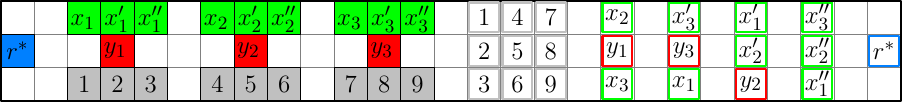}
        \caption{The instance $M'_3$ obtained from the instance $M(\P)$ where $\P$ is a Pivot Scheduling instance with the before constraints
        $X_1 = \set{x_1, x'_1, x''_1}, Y_1=\set{y_1}, X_2 = \set{x_2, x'_2, x''_2}, Y_2 = \set{y_2}, X_3 = \set{x_3, x'_3, x''_3}, Y_3 = \set{y_3}$
        and after constraints $\C=\set{\set{x_2,y_1,x_3}, \set{x'_3,y_3,x_1}, \set{x'_1, x'_2, y_2}, \set{x''_3,x''_2,x''_1}}$.} %
        \label{fig:no_obst_bounded}
    \end{figure}
}

\newcommand{\unbounded}[2]{
    \begin{figure}[#1]
        \centering
        \includegraphics[width=#2\textwidth]
        {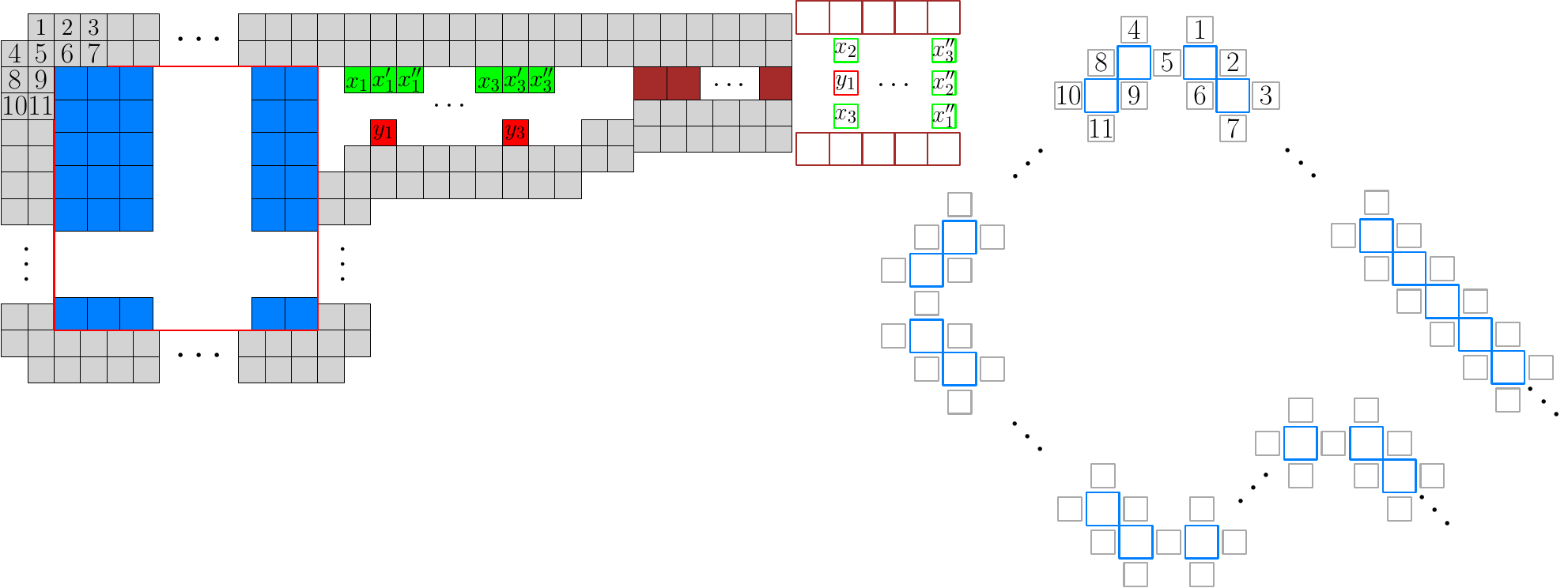}
        \caption{An example of $M'_4$ for $\varepsilon=1/4$ in
        the format of \Cref{fig:main-construction} except that gray and brown colors denote left and right boundary robots, respectively.
        For decluttering, we omit most robot labels; see text.}
        \label{fig:unbounded}
    \end{figure} 
}

\newcommand{\numagents}{\ensuremath{n}\xspace}
\newcommand{\numconstrs}{\ensuremath{N}\xspace}

\begin{document}

\maketitle

\begin{abstract}
We study the Monotone Sliding Reconfiguration (MSR)  problem, in which \emph{labeled} pairwise interior-disjoint objects in a planar workspace need to be brought \emph{one by one} from their initial positions to given target positions, without causing collisions.
That is, at each step only one object moves to its respective target, where it stays thereafter. %
MSR is a natural special variant of Multi-Robot Motion Planning (MRMP) and related reconfiguration problems, many of which are known to be computationally hard.
A key question is identifying the minimal mitigating assumptions that enable efficient algorithms for such problems.
We first show that despite the monotonicity requirement, MSR remains a computationally hard MRMP problem.
We then provide additional hardness results for MSR that rule out several natural assumptions.
For example, we show that MSR remains hard without obstacles in the workspace. %
On the positive side, we introduce a family of MSR instances that always have a solution
through a novel structural assumption pertaining to the graphs underlying the start and target configuration---we require that these graphs are
spannable by a forest of full binary trees (SFFBT). 
We use our assumption to obtain efficient MSR algorithms for unit discs and 2D grid settings. %
Notably, our assumption does not require separation between start/target positions, which is a standard requirement in efficient and complete MRMP algorithms.
Instead, we (implicitly) require separation between \emph{groups} of these positions, thereby pushing the boundary of efficiently solvable instances toward denser scenarios.

\end{abstract}

\section{Introduction}

We are given a set of labeled pairwise interior-disjoint objects (e.g., robots) %
in a polygonal workspace that need to be brought from their initial positions to given target positions, without causing collisions.
Such a setup appears in a host of geometric reconfiguration problems, such as Multi-Robot Motion Planning (MRMP), which vary by how a \emph{move} of an object is defined and additional constraints or objectives.
A \emph{sliding move}~\cite{DBLP:journals/comgeo/DumitrescuJ13} brings an object to another location in the plane using a continuous rigid motion.
In the \emph{sequential} case, only one object moves at a time, i.e., a solution is a sequence of single object moves.

\begin{figure}[t]
    \centering
    \begin{minipage}[b]{0.55\textwidth}
        \centering
        \includegraphics[width=1.02\textwidth]{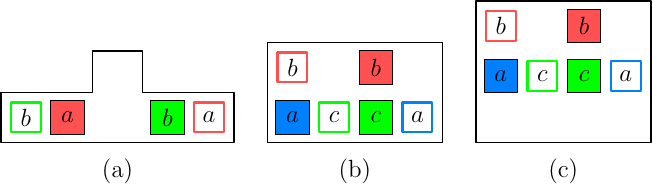}
        \caption{MSR instances. Filled (resp. unfilled) squares are start (resp. target) positions. (a)~A "No" instance, as one of the squares must make two moves. (b) An instance that can be solved by moving the squares in the order $b,a,c$ and no other order. (c) A modification of (b) to a well-formed instance (see text).}
        \label{fig:msr:intro}
    \end{minipage}
    \hfill
    \begin{minipage}[b]{0.42\textwidth}
        \centering
        \includegraphics[width=1.02\textwidth]{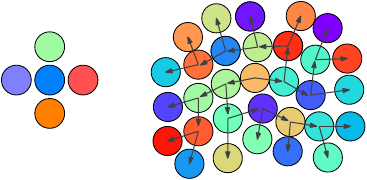}
        \caption{
        Two start configurations featuring disks with little to no separation between them, which are allowed under our assumption.
        The arrows (right) illustrate the core concept of SFFBT (see text). }
        \label{fig:intro:tight}
    \end{minipage}
\end{figure}

Given that reconfiguration problems are often computationally hard, identifying the least restrictive assumptions that enable solving them efficiently is a major quest.
Motivated by this question, we study a natural special case of sequential reconfiguration with sliding moves, which we call \emph{Monotone Sliding Reconfiguration (MSR)}.
MSR asks to find a \emph{monotone} solution, where each object moves at most once. See \Cref{fig:msr:intro}.
MSR appears to be the simplest variant in the long-studied aforementioned family of problems for which no efficient algorithm is known, thus motivating us to close this gap.
In turn, we perform a detailed investigation of the problem's computational complexity, which establishes its NP-hardness and rules out natural mitigating assumptions. On the positive side, we identify a key structural assumption that guarantees a solution.

\noindent\textbf{Background and related work.}
Mobile robots' continuously increasing role in manufacturing, logistics, and many other domains requires coordinating their motion safely and efficiently, giving rise to the MRMP problem.
In MRMP robots may move in parallel and make multiple moves, i.e., a robot stopping (to let another robot pass) and moving again is allowed. %
Already the feasibility version of  MRMP (where the motions need not be optimal) is known to be intractable in various planar settings~\cite{hopcroft1984complexity, DBLP:journals/ipl/SpirakisY84, DBLP:journals/tcs/HearnD05, DBLP:journals/ijrr/SoloveyH16, DBLP:conf/fun/BrockenHKLS21}.
Nevertheless pinpointing exactly when and why the problem is computationally hard, as a guide towards useful tractable variants, remains a major challenge.
For example, the complexity of MRMP for unit-disc robots is a longstanding open problem~\cite{DBLP:conf/fun/BrockenHKLS21}.
While for unit-square robots the problem is PSPACE-hard~\cite{DBLP:journals/ijrr/SoloveyH16}, the proof uses a polygonal workspace with holes and an unbounded complexity.\footnote{The workspace's complexity depends on the number of robots and cannot be expressed as a constant.}
A similar tradeoff occurs in a classic PSPACE-hardness result~\cite{hopcroft1984complexity}, which uses a rectangular workspace but relies on rectangular robots of various sizes. %
These examples highlight the challenge in understanding the complexity of MRMP when \emph{multiple facets}, such as robot shape and workspace shape, are \emph{simultaneously} simplified.
We strive towards this goal by considering MSR as a basic yet non-trivial MRMP variant, which we further restrict through additional assumptions.
Related work involving two of the authors shows that minimizing the total distance traveled by the robots is APX-hard in a weakly monotone setting~\cite{MRMP-RA-CG23} (see below) and NP-hard for robots on a 2D grid in the monotone (and non-monotone) case~\cite{DBLP:conf/atal/GeftH22}.
Both works fall short of directly addressing the issue of monotonicity, which is at the heart of the investigation of this work.

As MRMP hardness proofs often rely on tightly packed robots, complete%
\footnote{An algorithm is \emph{complete} if, in finite time, it returns a solution or reports that none exists.}
polynomial-time MRMP algorithms~\cite{DBLP:journals/tase/AdlerBHS15, DBLP:conf/rss/SoloveyYZH15, TangK15-spacing, yu-wafr-spacing, DBLP:journals/siamcomp/DemaineFKMS19, DBLP:conf/compgeom/BanyassadyBBBFH22} make \emph{separation assumptions}, such as a minimum distance between robots at their start or target positions. %
Another assumption for labeled unit-disc robots in a polygonal workspace~\cite{SolomonHalperin2018, MRMP-RA-CG23} requires each robot's start and target position to be contained in a \emph{revolving area}, i.e., a radius 
2 disc that is free from other robots' start and target positions.
The setting lends itself to \emph{weakly} monotone solutions, in which robots move one by one as in MSR, except that while one robot moves to its target, the other robots may move locally within their respective revolving areas. %
Separation is also implicitly required by a prevalent assumption, which is that an MRMP instance has a \emph{well-formed environment} (WFE)~\cite{DBLP:journals/tase/Cap0KS15} (see definition below).

Identifying the minimal separation that guarantees a solution that can be efficiently found is key for enabling robots to operate in crowded settings.
A recent work provides tight bounds for unlabeled%
\footnote{In the \emph{unlabeled} case any robot can go to any target, provided all the targets are eventually occupied.}
unit discs, showing that a distance of 4 between pairs of start positions and pairs of target positions is needed~\cite{DBLP:conf/compgeom/BanyassadyBBBFH22}.
Nevertheless, without such a separation a solution can still exist and the complexity of MRMP in such a case remains a challenging problem.
In particular, all the aforementioned assumptions exclude having just one robot tightly surrounded by others in the start or target configuration, such as the $5$ discs in \Cref{fig:intro:tight}~(left). %
Consequently, in this work we tackle the key question of which milder assumptions suffice to efficiently solve MRMP.

Related to MRMP is sequential reconfiguration in the plane, which has been investigated under various models and settings~\cite{DBLP:journals/comgeo/AbellanasBHORT06, DBLP:journals/siamdm/CalinescuDP08, DBLP:journals/comgeo/DumitrescuJ13, space-aware, DBLP:journals/ijcga/BeregDP08}.
Abellanas et al.~\cite{DBLP:journals/comgeo/AbellanasBHORT06} study the \emph{translation model}, in which a move is a single translation along a fixed direction, and provide lower and upper bounds on the required number of moves to reconfigure $n$ discs in various settings. %
They also give an $O(n^2)$-time algorithm for the special case of MSR under the translation model. %
Dumitrescu and Jiang~\cite{DBLP:journals/comgeo/DumitrescuJ13} show that minimizing the number of moves in the reconfiguration of (labeled or unlabeled) unit discs is NP-hard for either the translation or sliding model.
We use \emph{Reconfiguration with Minimum Number of Moves (RMNM)} to refer to the labeled case of their problem in our model. %
MSR is a special case of the latter requiring the least possible number of moves. %
The discrete version of RMNM, where objects are pebbles on a graph, was shown to be NP-hard on the 2D grid~\cite{DBLP:journals/siamdm/CalinescuDP08} and W[1]-hard on general graphs~\cite{DBLP:conf/walcom/CooperMMN22}. 
For more related work see references within~\cite{space-aware, mover-probs}.

Similar problems arise as \emph{rearrangement} problems in robotics, e.g., when a robotic arm needs to rearrange products on a shelf, where the combinatorial problem nature results in various challenges~\cite{DBLP:conf/icra/GaoLHBY22, DBLP:journals/ijrr/GaoFHY23,DBLP:conf/icra/GaoY23}.
Minimizing the number of moves shortens the time needed to perform such tasks, e.g., through fewer pick and place operations. 
In this vein, MSR for unit discs has been recently tackled empirically, where the problem was shown to be challenging already for 10-30 discs~\cite{wang2021uniform}, thus further motivating the study of MSR. %

\vspace{-6pt}
\subsection*{Contribution}
\vspace{-2pt}
\textbf{Negative results.}
First, we provide a simple construction that establishes the NP-hardness of MSR.
We then present stronger hardness results through additional restrictions along two facets.
In the first facet (see \Cref{sec:relaxing}), we examine two previously studied tractable MSR variants and show that an immediate generalization of each of them becomes intractable:
In a \emph{well-formed environment (WFE)}~\cite{DBLP:journals/tase/Cap0KS15} every robot has a path from its start to its target position that is \emph{endpoint-free}, which is a path that does not intersect any other robot's start or target position.
The assumption is strong in the sense that under it the robots may move to their targets one by one in \emph{any} order.
We show that for the slightly milder \emph{nearly well-formed environment (nearly WFE)}, where only one robot does not have an endpoint-free path, MSR is NP-hard.
Next, for MSR under the translation model~\cite{DBLP:journals/comgeo/AbellanasBHORT06} we show that if only one designated robot is allowed a sliding move instead of a single translation, then the problem becomes NP-hard as well.

The second facet of our results (see \Cref{sec:hardness-no-obst}) tackles a key element of previous MRMP and reconfiguration hardness proofs--obstacles in the workspace.
We show that MSR remains hard without obstacles in conjunction with another assumption:
We require that the start and target configurations are \emph{line separable}, i.e., there exists a vertical line such that all start positions are left of it and all target positions are right of it. 
Line separability prohibits robots from emulating obstacles by staying in place (which would give a trivial negative statement).
Here we distinguish between a \emph{bounded} workspace (in which the robots are confined) and an \emph{unbounded} workspace (allowing robots to move anywhere in the plane).
We consider the negative result for the latter case to be our strongest negative result as this setting is significantly less constrained and hence requires the most intricate construction.
\Cref{thm:hardness} summarizes the results.

These results point to a scheduling problem called Pivot Scheduling~\cite{fixed-path} as the underlying source of the intractability of all studied variants. We therefore turn towards a structural assumption that avoids the hardness of Pivot Scheduling. %

\noindent\textbf{Positive results.}
For the discrete version of MSR, we define a family of pebble motion graphs (PMGs) that always have a solution through a novel structural assumption.
We show that if a PMG is \emph{spannable by a forest of full binary trees (\shortsolvable{})}, or \emph{spannable} for short, (see \Cref{sec:positive:assumption} for details) then the MSR instance has a solution that can be efficiently found given the forest.
Next, we turn to the problem of recognizing spannable PMGs, i.e., deciding whether a discrete MSR instance belongs in our family.
Although we show that this problem is NP-hard for general graphs, we obtain positive results for geometric MSR instances.
Specifically, for unit-disc robots we show that we can efficiently compute a corresponding PMG and decide whether it is spannable. 
We obtain this result by showing that in this case spannable PMGs have constant treewidth, which allows applying Courcelle's theorem~\cite{DBLP:journals/iandc/Courcelle90} (a brief background is given in \Cref{sec:positive:spannability}).
We also prove that a special variant of MSR for grid-aligned unit squares in which the start and target configurations induce a set of simple thin polyominoes (defined in \Cref{sec:positive:polyomino}) always has a spannable PMG.

Notably, through our structural assumption we show that a solution can be guaranteed and efficiently found under a relaxation of standard separation assumptions used in complete and efficient MRMP algorithms.
In our setting, we replace separation between individual (non-overlapping) start/target positions with a requirement for separation between "groups" of such positions.
A potential consequence is enabling more space-efficient object reconfiguration~\cite{space-aware} and deployment of robot teams. %
This work can also be seen as a step towards formal workspace design~\cite{DBLP:conf/atal/SalzmanS20}, where guaranteeing solutions is crucial, in higher density.

\smallskip
\noindent
\textbf{Notation and terminology.}
In MSR we are given a set $R$ of $n$ robots (or objects), which are either all axis-parallel squares or all unit discs.
The robots operate in a planar, possibly polygonal, workspace, where each $r \in R$ has a start and target position, denoted by \src{r} and \trg{r}, respectively, which we also call \emph{endpoints}.
A \textit{solution} to MSR is a \emph{monotone motion plan}, which is a sequence of moves in which each robot $r \in R$ performs a translational motion \emph{once} from \src{r} to \trg{r} along a collision-free path $\pi(r)$ while all other robots are stationary.
If such a solution exists for an MSR instance we call it \emph{feasible} or \emph{solvable}.
We distinguish between two variants with regard to the robots' uniformity:
In \MSRsq $R$ and the obstacles consist of grid-aligned unit squares, which is equivalent to the discrete case of pebbles moving on the 2D grid or grid-aligned unit discs.
In
\MSRnear $R$ consists of \emph{nearly identical squares}, i.e., each square has a side length of $1$ or $1 + \varepsilon$, for a fixed arbitrarily small $\varepsilon > 0$.

\vspace{-4pt}
\section{Hardness results}\label{sec:hardness}

In this section, we establish the following hardness results:
\begin{theorem}
\label{thm:hardness}
MSR is NP-hard in each of the following variants:
\begin{enumerate}[label=(\roman*), align=left]
    \item Nearly WFE variant of \MSRnear, %
    \item \MSRsq where all the squares except one must move via a single axis-parallel translation,
\end{enumerate}
and the following obstacle-free line-separable variants:
\begin{enumerate}[label=(\roman*), resume, align=left]
    \item \MSRsq for a $3 \times \ell$ rectangular workspace,
    \item \MSRnear for an unbounded workspace.
\end{enumerate}
\end{theorem}

We first establish the hardness of MSR (without additional assumptions) using a reduction from a variant of Pivot Scheduling~\cite{fixed-path}, which obtains an MSR instance $M$.
Then, for each MSR variant in \Cref{thm:hardness}, we show how to convert $M$ to an instance $M'$ of the variant such that $M$ is solvable if and only if $M'$ is solvable.

\subsection{Establishing the hardness of MSR} \label{sec:basic-hardness}
Let us define the base problem for our reduction.

\noindent
\textbf{Pivot Scheduling.} %
Let $V$ be a set of jobs.
Let $\X = \set{X_1, \ldots, X_n}$ and $\Y = \set{Y_1, \ldots, Y_n}$ be $2n$ pairwise disjoint subsets of $V$, i.e., $X_i, Y_i \subseteq V$ for each $i$.
Let $\C = \{C_1, \ldots, C_m\}$, where $C_j \in V^3$ for each $j$, and the $C_j$'s are pairwise disjoint.
The problem is to determine whether $V$ can be partitioned into a before-set $\bef$ and an after-set $\aft$ (i.e., $\aft \cap \bef = \emptyset$ and $\aft \cup \bef = V$) under the following constraints.
\emph{Before constraints:} For each $i \in [n]$, either $X_i \subseteq \bef$ or $Y_i \subseteq \bef$.
\emph{After constraints:} For each $j \in [m]$, we have $C_j \cap \aft \neq \emptyset$.

\begin{figure}[t]
\centering
\includegraphics[width=0.55\textwidth]{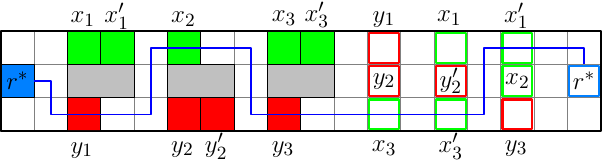}
\caption{The MSR instance $M$ for the Pivot Scheduling instance with the before constraints
$X_1 = \set{x_1, x'_1}, Y_1=\set{y_1}, X_2 = \set{x_2}, Y_2 = \set{y_2, y'_2}, X_3 = \set{x_3, x'_3}, Y_3 = \set{y_3}$
and after constraints $\C=\set{\set{y_1,y_2,x_3}, \set{x_1,y'_2,x'_3}, \set{x'_1,x_2,y_3}}$.
Obstacles appear in gray.
The start and target positions are the filled and unfilled colored squares, respectively.
The start and target positions of the $R(X_i)$ (resp. $R(Y_i)$) robots are green (resp. red).
Labels distinguish between robots having the same color.
The path \ppath{} (blue) is shown for the solution $\bef{} = \set{y_1, x_2, y_3}, \aft{} = \set{x_1,x'_1, y_2,y'_2, x_3,x'_3}$.
}
\label{fig:main-construction}
\end{figure}

One can think of each constraint as being imposed by an implicit \emph{pivot job} where the jobs of \bef{} are to be executed before this pivot job and the jobs of \aft{} are to be executed after it.
For an example instance and a solution see %
\Cref{fig:main-construction}.
A similar version of the problem is known to be NP-hard~\cite{fixed-path}. Our formulation differs slightly for convenience. We prove its hardness using a reduction from a special 3SAT variant~\cite{DBLP:journals/dam/DarmannD21} (see \Cref{sec:missing}).

\begin{restatable}{lemma}{pivot} 
\label{thm:pivot-sched}
Pivot Scheduling is NP-hard even when $|X_i| =3$, $|Y_i|=1$ for all $i$.
\end{restatable}

\noindent
\textbf{A unified approach for proving the hardness of MSR variants.}
The common elements of our proofs are as follows.
We represent each job by a robot and additionally use a distinguished \emph{pivot robot} $r^*$ (or multiple such robots) that emulates the implicit pivot job in Pivot Scheduling.
The robot $r^*$ starts at the left side of the workspace %
and then has to proceed rightward through \emph{gadgets} that represent Pivot Scheduling constraints.
First, $r^*$ proceeds through \emph{before-constraint} gadgets.
For such a gadget, which we denote by $B_i$, $r^*$ must choose between one of two paths to traverse the gadget.
Each path is initially blocked by robots, which have to move before $r^*$.
Robots on one path correspond to $X_i$ while those on the other path correspond to $ Y_i$.
After proceeding through all before-constraint gadgets, $r^*$ has to proceed through after-constraint gadgets.
Each after-constraint gadget, denoted by $A_j$, can be traversed using one of three paths, each passing through a target of a robot representing a job in $C_j$.
Therefore, for $r^*$ to traverse the gadget, one of the paths has to have an unoccupied target, i.e., $A_j$ enforces that one of the jobs in $C_j$ appears in the after-set $\aft{}$.

\noindent
\textbf{Hardness of MSR.}
Given a Pivot Scheduling instance $\P=(V, \C)$, we construct a MSR instance $M \coloneqq M(\P)$ that is feasible if and only if $\P$ is feasible.
The workspace of $M$ is a three-row high rectangular portion of the unit grid; see Figure~\ref{fig:main-construction}.
All the start and target positions, as well as obstacles, are grid cells.
There is a robot in $M$ for each job in $V$ and a distinguished pivot robot $r^*$.
For a set of jobs $V' \subseteq V$, we denote by $R(V')$ the robots that correspond to the jobs in $V'$.
All the non-pivot robots' start positions are located at before-constraint gadgets.
For each pair $X_i, Y_i$ in $\P$ the gadget $B_i$ has obstacle cells in its middle row, which form a top (resp. bottom) path initially containing the robots $R(X_i)$ (resp. $R(Y_i)$).
The $B_i$'s are separated from each other by columns of empty cells.

The non-pivot robots' target positions are located in after-constraint gadgets, each corresponding to an after constraint $C_j$.
The gadget consists of a column of three cells, which are the targets of $R(C_j)$.
The after-constraint gadgets are separated from each other by columns of empty cells.
Finally, $r^*$ must traverse the workspace from left to right passing through all gadgets.
The order of gadgets of the same type is arbitrary but determines the order of the start positions:
The left to right order of start positions within each before-constraint gadget is set to match the left to right order of the corresponding target positions.
We refer to this ordering as the \emph{start-target ordering correspondence}.

\begin{theorem} \label{thm:monotone-MRMP}
MSR is NP-complete.
\end{theorem}

\begin{proof}
Given a motion plan for $M$, let $R_1$ and $R_2$ denote the robots that move before and after $r^*$ in the motion plan.
We have a correspondence between valid job partitions $(\bef, \aft)$ and $(R_1, R_2)$ pairs, which describe motion plans for $M$.
Given a valid motion plan, the corresponding partition of $V$ satisfies all the constraints:
Each before-constraint is satisfied since either all the robots in $R(X_i)$ or all the robots in $R(Y_i)$ must move so that $r^*$ can traverse $B_i$ to its target.
Similarly, each after-constraint is satisfied since one of the targets of each $A_j$ must be unoccupied by a robot for $r^*$ to traverse the gadget.

For the other direction, we show that if $(\bef, \aft)$ is a valid job partition, the robots can move in the order $R(\bef{}), r^*, R(\aft{})$.
We first define the path for $r^*$, which we refer to as the \textbf{partition path} and denote by $\ppath{}$.
We take $\ppath{}$ to be the shortest collision-free path from \src{r^*} to \trg{r^*} that passes through start positions of the robots in $R(\bef{})$ and targets of the robots in $R(\aft{})$ (and no other endpoints).
\ppath{} exists in $M$ since $(\bef{}, \aft{})$ is valid, meaning that each gadget contains the start/target positions that enable $\ppath{}$ to traverse it. %

The robots in $R(\bef{})$, which are initially located on \ppath{}, move in the right to left order of their start positions as follows.
When a robot $r \in R(\bef{})$ moves, it moves along \ppath{} until it reaches the after-gadget that contains its target position, at which point the robot strays off \ppath{} to reach its target.
Due to the order in which $R(\bef{})$ move, when $r$ moves any starting position through which it passes is unoccupied.
Any target that $r$ passes through is also unoccupied, since the robots of $R(\aft{})$ have not moved at this stage.
Therefore, $r$'s motion is collision-free.
Next, $r^*$ moves to its target using \ppath{}.
At this point, all the start and target positions that lie on \ppath{} are unoccupied, so this move is collision-free.
Finally, the robots of $R(\aft{})$ move in the right to left order of the target positions, similarly to robots of $R(\bef{})$.
Due to the order in which $R(\aft{})$ move, when some $r \in R(\aft{})$ moves any target position through which it passes is unoccupied.
The start-target ordering correspondence ensures that any start position that $r$ passes through is also unoccupied.
Therefore, $r$'s motion is collision-free and so overall we described a valid motion plan.
Lastly, NP membership follows from standard techniques for translational motion planning in the plane~\cite[Chapter~13]{Berg:2008:CGA:1370949}.

\end{proof}

\subsection{Relaxing assumptions of tractable cases}\label{sec:relaxing}
In this section, we examine two previously studied MSR variants that are polynomial-time solvable.
We show that a slight relaxation of the assumptions in these variants results in making the problem NP-hard, thereby proving \Cref{thm:hardness}~(i) and (ii).

\noindent\textbf{Nearly well-formed environments.}
\label{sec:WFe} %
One approach for tackling the hardness of MRMP is to introduce simplifying assumptions on the workspace.
C{\'{a}}p et al.~\cite{DBLP:journals/tase/Cap0KS15} introduced the notion of a \emph{well-formed environment}, in which a robot located at its start or target position cannot block other robots.
Formally, for every robot $r$, there exists a path from $\src{r}$ to $\trg{r}$ that does not intersect any other source or target.
A robot satisfying this property is said to have an \emph{endpoint-free path}. %
It is easily seen that under this assumption a monotone motion plan exists for any ordering of the robots.
A natural question is whether we need all robots to have the property above.
We show that for a \emph{nearly well-formed environment (nearly WFE)}, where the property does not hold for only one robot, the problem is already hard.

\begin{figure}[t]
\centering
\includegraphics[width=0.54\textwidth]{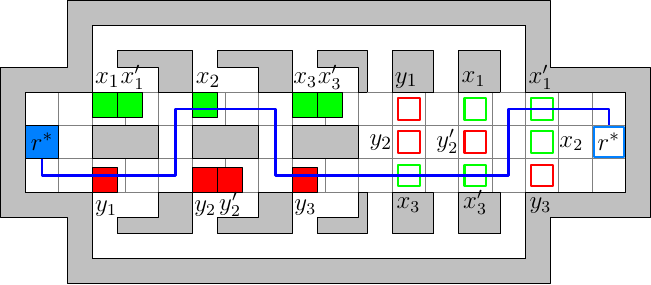}
\caption{The instance $M'_1$, obtained from the instance $M$ (\Cref{fig:main-construction}), for $\varepsilon = 1/3$.} %
\label{fig:nearly-WFE}
\end{figure}

We now adapt the instance $M$ from Theorem~\ref{thm:monotone-MRMP} to an instance $M'_1$ that constitutes a nearly WFE.
Let $\R$ denote the three-row high grid-aligned rectangle that constitutes the workspace of $M$.
We use $\R$ to refer to the same portion of the workspace in the new instance $M'_1$.
The main idea is to add two paths, one above $\R$ and one below $\R$, that allow non-pivot robots to reach their targets without passing through any gadgets; See \Cref{fig:nearly-WFE}.
The new paths can be accessed from each non-pivot robot's start position via unit-width passages.
As a result each non-pivot robot has an endpoint-free path to its target.
On the other hand, $r^*$ now has a side length of $1 + \varepsilon$, which prohibits it from accessing the new paths thus enforcing its passage through all the gadgets.

We now give more precise details on the transformation of $M$ into $M'_1$.
We first uniformly scale up $M$ so that the grid cells have a side length of $1 + \varepsilon$.
We do so without scaling up the non-pivot robots, which thus remain unit squares.
The start positions of robots in each $R(X_i)$ (resp. $R(Y_i)$) are moved slightly to be aligned along $\R$'s top (resp. bottom) edge.
Each gadget has two vertical unit-width passages, one for accessing the new top path and one for accessing the new bottom path.
To enable a robot in $R(X_i)$ to reach the top path, we add some space that allows it to translate up by exactly one unit, then left to reach the unit width passage above $B_i$; $R(Y_i)$'s are handled symmetrically. 
Observe that the space added above/below the start positions in each $B_i$ does not allow $r^*$ to fully leave $\R$ at any time.
Consequently, we still have the property that for each $B_i$ either all of $R(X_i)$ or all of $R(Y_i)$ must move before $r^*$ can traverse $B_i$.

We now establish the correctness of the reduction.

\begin{restatable}{lemma}{nearlyWFE} \label{thm:nearly-WFE}
$M'_1$ is feasible if and only if $M$ is feasible.
\end{restatable}

\begin{proof}
Suppose $M'_1$ is feasible and let us examine $\pi(r^*)$ in the solution to $M'_1$.
Without loss of generality, we assume that $\pi(r^*)$ is weakly $x$-monotone.
The robots along $\pi(r^*)$ may be moved in right to left order of their start positions in $M$.
This clears a path for $r^*$ to reach its target in $M$, after which we may move the rest of the robots to their targets (see proof of \Cref{thm:monotone-MRMP}).

For the other direction, suppose $M$ is feasible.
We move the robots in $M'_1$ in the same order as in the solution to $M$.
This is valid since each non-pivot robot can reach its target via an endpoint-free path regardless of when it moves.
As for $r^*$, it can move using the partition path \ppath{}, which exists and is free of robots during $r^*$'s motion. %
\end{proof}

\noindent
\textbf{All but one of the robots have a fixed path.}
Another way in which MRMP can be restricted to be easier to solve is by fixing the paths of the robots.
That is, as part of the input we also get the path that each robot must move along to reach its target.
In this case, MSR can be solved in polynomial time by means of a precedence graph~\cite{DBLP:conf/icra/Buckley89, DBLP:journals/comgeo/AbellanasBHORT06}.

We highlight another tractability frontier by showing that it suffices to only have one robot with an unspecified path to make the problem hard, i.e., MSR where every robot must follow a given path except for one robot.
Note that instance $M'_1$ above can be readily used to show this, i.e., assign each non-pivot robot a path that does not interfere with other robots.
However, we wish to prove the stronger requirement of \Cref{thm:hardness}~(ii), where every given path is an axis-parallel translation, which requires a more involved construction.

\begin{figure}[htb]
\centering
\includegraphics[width=0.55\textwidth]{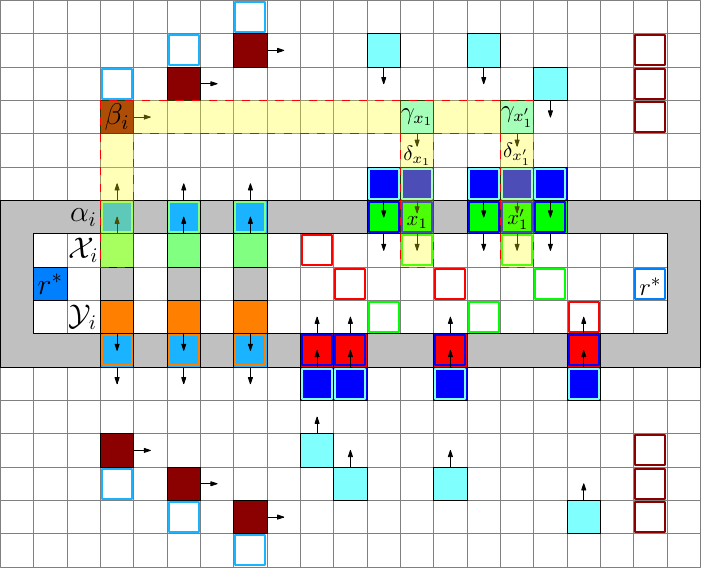}
\caption{
The instance $M'_2$, obtained from the instance $M$ (\Cref{fig:main-construction}).
For clarity, we label the robots associated with $X_i$, which are highlighted using yellow-shaded cells, and omit most other labels.}
\label{fig:nearly-fixed-path}
\end{figure}

To this end, we modify $M$ (used to prove Theorem~\ref{thm:monotone-MRMP}) to an instance $M'_2$ with the required properties.
Specifically, the robots and obstacles remain grid-aligned unit squares and all non-pivot robots have to make a single axis-parallel translation.
We use additional grid rows above and below $M$, where new auxiliary robots will move.
Since all the fixed paths are line segments, the endpoints of each of the non-pivot robots define the paths they must follow.
Figure~\ref{fig:nearly-fixed-path} shows an example of $M'_2$.
To help read it, each non-pivot robot has an arrow that indicates the direction of its path.
Each such robot's target is the first one encountered in the arrow's direction. %

We now describe the modifications to the gadgets.
Let $\R$ denote the three-row high rectangular portion of the grid that constitutes the workspace of $M$, which we now modify as follows.
After-gadgets are modified so that the targets form a diagonal on a $3\times3$ grid portion of $\R$, i.e., each target in $\R$ lies on a unique column.
All the $R(X_i)$'s (resp. $R(Y_i)$'s) are initially located one row above (resp. below) $\R$ such that each non-pivot robot is located in the same column as its target.

Each before-gadget $B_i$ now has two new robots: $\X_i$ at its top row and $\Y_i$ at is bottom row (while the middle row has an obstacle cell).
We add auxiliary robots that ensure that $\X_i$ (resp. $\Y_i$) can move only after all the robots of $R(X_i)$ (resp. $R(Y_i)$) have moved.
We now describe how to enforce such precedence constraints for some $X_i$ (the auxiliary robots for $Y_i$ act symmetrically).

For each $X_i$ set, we have the following additional new robots in $M'_2$: $\alpha_i$, $\beta_i$, and two robots for each job in $X_i$.
We collectively refer to these new robots and $\X_i$ as the \emph{auxiliary robots} of $X_i$.
We describe the relative placements of these robots and then give exact coordinates.
The cell above \src{\X_i} is both \trg{\X_i} and \src{\alpha_i}.
We place \trg{\alpha_i} $i+3$ cells above \src{\alpha_i} and place \src{\beta_i} a cell below \trg{\alpha_i} (i.e., \src{\beta_i} lies on $\alpha_i$'s path).
We place \trg{\beta_i} in the rightmost column of $\R$.
These placements ensure that $\beta_i$ must move before $\alpha_i$, which in turn must move before $\X_i$.
For each robot $r \in R(X_i)$, we have two additional auxiliary robots, $\gamma_r$ and $\delta_r$, which have to go down along the column in which $r$ lies.
We place \src{\gamma_r} on $\beta_i$'s path, with \trg{\gamma_r} being the same cell as \src{\delta_r}, and \trg{\delta_r} being the same cell as \src{r}.
These placements ensure that $r$ moves before $\delta_r$, which must move before $\gamma_r$, which in turn must move before $\beta_i$.
Combining all the aforementioned precedence constraints, we indeed have the constraint that all the robots in $R(X_i)$ must move before $\X_i$ does (see also \Cref{fig:nearly-fixed-path}).

We now define the exact endpoints of the auxiliary robots for a set $X_i$ and a robot $r \in X_i$.
Let us set the coordinates of the top left cell in $\R$ to be $(0,0)$. %
Also, let $c_1$ denote the $x$-value of the column containing $r$ and $c_2$ denote the $x$-value of the rightmost column of $\R$. %
We define the endpoints as follows:
\begin{itemize}[label={}]
    \item $\src{\X_i} \coloneqq (2i, 0)$,
    \item $\trg{\X_i} = \src{\alpha_i} \coloneqq (2i, 1)$,
    \item $\trg{\alpha_i} \coloneqq (2i, i+4)$,
    \item $\src{\beta_i} \coloneqq (2i, i+3)$,

    \item $\trg{\beta_i} \coloneqq (c_2, i+3)$,
    
    \item $\src{\gamma_r} \coloneqq (c_1, i+3)$,
     
    \item $\trg{\gamma_r} = \src{\delta_r} \coloneqq (c_1, 2)$,
     
    \item $\trg{\delta_r} \coloneqq \src{r} = (c_1, 1)$.
    
\end{itemize}
Note that the $y$-coordinate of $\src{\beta_i}$ depends on $i$ so that $\beta_i$ can move rightward along a row that is not used by auxiliary robots of other $X_i$'s.
More precisely, the placement of the endpoints results in paths such that the auxiliary robots of an $X_i$ can move at any stage without regard to other auxiliary robots.

Finally, each cell adjacent to $\R$ and outside of it is an obstacle, except for the cells that are start/target positions.
We let $r^*$ be the robot whose path is not given.
The obstacles surrounding $\R$ are designed such that $r^*$ cannot leave $\R$ and thus its motion is similar to the motion of the corresponding robot in $M$, which we state formally in the following lemma:

\begin{lemma} \label{thm:fixed-path}
$M'_2$ is feasible if and only if $M$ is feasible.
\end{lemma}
\begin{proof}
Let us assume that $M'_2$ has a monotone motion plan.
It is easy to verify that $r^*$ cannot leave $\R$ during its motion.
Therefore, in each before-gadget $B_i$, either $\X_i$ or $\Y_i$ must move before $r^*$.
Consequently, by construction either all of $R(X_i)$ or all of $R(Y_i)$ must move before $r^*$ traverses the before-gadget $B_i$.
For each after-gadget $A_j$, one of the robots whose target is in the gadget must move after $r^*$.
Let us examine $\pi(r^*)$, which is weakly $x$-monotone without loss of generality. %
We move the following robots in $M$ in right to left order of each $B_i$: we move $R(X_i)$ if $r^*$'s path goes through $\src{\X_i}$ and $R(Y_i)$ if $r^*$'s path goes through $\src{\Y_i}$.
It is readily seen that this order is valid in $M$ and that it clears a path for $r^*$ in $M$, after which the rest of the robots can move to their targets (see proof of \Cref{thm:monotone-MRMP}).

For the other direction, let $R_1$ and $R_2$ denote the robots that move before and after $r^*$ in a solution to $M$.
We solve $M'_2$ by moving robots corresponding to $R_1$ and their respective auxiliary robots, followed by $r^*$, and then robots corresponding to $R_2$ and their respective auxiliary robots.

Let $R'(X_i)$ (resp. $R'(Y_i)$) denote $R(X_i)$ (resp. $R(Y_i)$) and the auxiliary robots of $X_i$ (resp. $Y_i$).
Note that without any loss of generality, each $R(X_i)$/$R(Y_i)$ in $M$ is contained in either $R_1$ or $R_2$.
Therefore, the robots in each $R'(X_i)$/$R'(Y_i)$ set move consecutively in the resulting motion plan for $M'_2$. %
The resulting motion plan is valid since the motion of each robot set $R'(X_i)$/$R'(Y_i)$ does not depend on any robot not in the set and can always be performed (using an appropriate ordering).
As for $r^*$, it can move using a modified version of the partition path \ppath{}: %
If originally \ppath{} passes through the start positions of $R(X_i)$ (resp. $R(Y_i)$), it now passes through $\src{\X_i}$ (resp. $\src{\Y_i}$).
It is easily seen that the modified path is free of robots during $r^*$'s motion.
\end{proof}

\subsection{Hardness without obstacles}\label{sec:hardness-no-obst} %
In this section we prove the hardness of MSR when there are no obstacles in the workspace.
We assume here that $M$ (the instance in Theorem~\ref{thm:monotone-MRMP}) has $|R(X_i)|=3$ and $|R(Y_i)|=1$ in each $B_i$, which is valid due to \Cref{thm:pivot-sched}.

\medskip
\subsubsection*{Bounded workspace}%
\label{sec:no-obst-bounded} %
We show that \MSRsq remains hard in a rectangular workspace without obstacles.
We describe how to adapt $M$ to an instance $M'_3$ with the aforementioned properties.
The idea is to modify the before gadgets in $M$ so that obstacles are now \emph{obstacle robots}.
To make room for the targets of the new robots, we insert columns to the grid to the right of before gadgets. %
Each $B_i$ now occupies a $3\times3$ subgrid in which the top row contains $R(X_i)$, the middle cell in the middle row contains the single robot in $R(Y_i)$, and the bottom row contains three obstacle robots.
The targets of the three obstacle robots are placed in a single column in one of the newly inserted columns; see Figure~\ref{fig:no_obst_bounded}.
We note that the order of such columns with respect to each other can be arbitrary (but each column contains targets corresponding to a single before gadget).

\ifthenelse{\boolean{ispaper}}{ %
}{ %
\noObstBounded{htb}{0.75} %
}

\Needspace{3\baselineskip}
\begin{lemma} %
$M'_3$ is feasible if and only if $M$ is feasible.
\end{lemma}
\begin{proof}
Assume that $M'_3$ has a monotone motion plan.
Observe that $r^*$ may only traverse $B_i$ by passing through the start positions of $R(X_i)$ or the start position of $R(Y_i)$.
The only other option is passing through all three start positions of obstacle robots.
However, if the three obstacle robots move before $r^*$, they will form an untraversable obstacle at their corresponding target positions.
Therefore, a solution to $M'_3$ emulates as solution for $M$.

For the other direction, we move the robots in the same order as in a solution to $M$, which is clearly possible.
Then, we move obstacle robots in right to left order of their targets.
\end{proof}

\medskip
\subsubsection*{Unbounded workspace} \label{sec:no-obst-unbounded}\noindent We now show that MSR is NP-hard for nearly identical squares in an unbounded workspace with no obstacles using an instance $M'_4$. 
Similarly to the previous reduction here we also emulate obstacles with robots, with the difference that now they emulate boundaries in the workspace.
We use \emph{left (resp. right) boundary} to refer to the boundary that partially encloses the start (resp. target) positions.
The left (resp. right) boundary is formed by robots located at their start (resp. target) positions, which we call \emph{boundary robots}.
For this reduction we use multiple pivot robots.
All the pivot robots and right boundary robots have a $1 + \varepsilon$ side length; the rest of the robots are unit squares.
The exact number of robots and overall dimensions of $M'_4$ are made more precise in the sequel.

\noindent
\textbf{Left boundary.} 
We now give an overview of the part of the construction containing start positions.
Refer to \Cref{fig:unbounded}.
This part is partially enclosed by the \emph{left boundary}, which forms a \emph{square room} (large red square in \Cref{fig:unbounded}), where the pivot robots initially reside.
The square room is connected to a narrow corridor, called the \emph{left corridor}, on its right, which leads to another corridor formed by the right boundary, called the \emph{right corridor}.
The left boundary is a two-layer thick wall of unit squares, which are the start positions of boundary robots (colored gray).
These unit squares are grid-aligned except for a section where the corridor narrows (to be shortly addressed).
For example, the top portion of the boundary occupies a $2 \times \ell$ rectangular portion of the unit grid, where each cell is a start position.
For each corner of the boundary that is adjacent to the aforementioned square room
(there are four such corners), we do not have a start position on the grid square on the outer side of the corner. %

\ifthenelse{\boolean{ispaper}}{ %
}{ %
    \unbounded{htb}{1} %
}

The left corridor initially has a height of $3$.
This portion of the corridor contains the before-constraint gadgets, which are similar to the ones in $M'_3$ (\Cref{fig:no_obst_bounded}).
Each gadget is $3\times3$ portion of the grid, with the top row containing $R(X_i)$'s and the bottom containing $R(Y_i)$.
The gadgets are placed one unit apart.
After the before-gadgets, the corridor narrows to a height of $1 + \varepsilon$.
This narrower portion contains the start positions of right boundary robots (colored brown).
The corridor's narrowing occurs using two \emph{upward shifts}, as shown.
First, we have a $2 \times 2$ block (consisting of start positions) shift up by one unit with respect to the previous such block to its left.
Then, the next block is shifted up similarly by $1-\varepsilon$, and the following blocks comprising the boundary on the right are at the same height.

\noindent
\textbf{Mapping the left boundary to target positions.}
We describe the target configuration of the left boundary robots using a bijective transformation of their start positions.
For this purpose, we ignore the aforementioned upward shifts that make the left corridor narrower, i.e., we let the whole left corridor have a height of $3$ here.
The transformation takes the centers of the start positions and then rigidly rotates them by $45^{\circ}$ clockwise.
We then uniformly scale the rotated points so that they are farther apart by a factor of $(2 + \varepsilon)/\sqrt{2}$.
Target positions are then placed so their centers lie on the resulting points.
The resulting configuration is placed in the unbounded area of the workspace, i.e., not enclosed by any boundary.
The labels in~\Cref{fig:unbounded} illustrate the mapping of the top-left portion of the boundary to corresponding target positions.
Notice that any $2\times2$ block of start positions corresponds to a "diamond" of four target squares (i.e., the four squares' centers lie on vertices of a larger square rotated by $45^{\circ}$).
We use this fact to enforce that the left boundary is in place when the pivot robots move.
We do so by placing the target of each pivot robot inside of an (arbitrary) aforementioned diamond.

The right corridor has a height of $3(1 +\varepsilon)$ and is formed by two horizontal walls consisting of targets of right boundary robots (arbitrarily mapped from start positions).
The corridor contains after gadgets spaced $1 + \varepsilon$ units apart.
The right opening of the right corridor leads to the unbounded area of the workspace.
The corridor's left opening is placed $\varepsilon/2$ units away from the opening of the left corridor, which ensures that pivot robots can only reach the unbounded area of the workspace through the right corridor.

\noindent
\textbf{Size of the construction.}
We now show that $M'_4$ has a size linear in the size of $M$.
Let \numconstrs denote the number of after constraint gadgets (which is the same in $M$ and $M'_4$).
The number of before constraint gadgets is $3\numconstrs/4$ and it suffices to have $4\numconstrs$ right boundary robots.
Therefore, the total length required for the left corridor is $O(\numconstrs)$.
Let $\numconstrs'$ denote the number of left boundary robots needed for the left corridor, which is also $O(\numconstrs)$.
We set the side length of the square room enclosed by the left boundary to be $\numconstrs'$.
To achieve this side length, we need at most $8\numconstrs'$ additional left boundary robots, bringing their total number to $9\numconstrs'$.
Hence, the number of pivot robots required is at most $9/2\numconstrs'$, and therefore the total number of robots is $O(\numconstrs)$.
The area available for placing the start positions of the pivot robots is $\numconstrs'^2$, which suffices for a sufficiently small $\varepsilon$.

\begin{restatable}{lemma}{unboundedlem}
$M'_4$ is feasible if and only if $M$ is feasible.
\end{restatable}

\begin{proof}
    Let us assume that $M'_4$ has a motion plan and let $r_1^*$ denote the first pivot robot that moves in it.
    We claim that for each $B_i$, $r_1^*$ may only traverse $B_i$ by passing through the start positions of $R(X_i)$ or the start position of $R(Y_i)$.
    Observe that the only other possibility involves first moving some $2 \times 2$ block of robots on the left boundary. %
    However, if four such robots move before $r_1^*$ they would block off a target of a pivot robot, which is a contradiction.
    By the same argument, $r_1^*$ has to pass through all the start positions of right boundary robots.
    Hence, the boundary of the right corridor must be in place before $r_1^*$ moves.
    Therefore, $r_1^*$ must pass through one of the three target positions in each after constraint gadget $A_j$.
    The constraints imposed on $r_1^*$'s motion are the same as those imposed on $r^*$ in $M$.
    Therefore we may obtain a motion for $M$ by moving the robots that appear on $\pi(r_1^*)$ in right to left order of their starting positions in $M$ and then move $r^*$. %
    Therefore, a solution to $M'_4$ emulates a solution for $M$.
    
    For the other direction, let us assume that $M$ has a monotone motion plan.
    We solve $M'_4$ by first moving right boundary robots (brown) in right to left order of their start positions.
    Then we move robots according to the solution of $M$ with the following adaptation:
    When is it the $r^*$'s turn to move, we move all the pivot robots (blue) in $M'_4$ in top-down, right to left order of their start positions.
    Finally, we move the left boundary robots (gray), which can be done in arbitrary order.
\end{proof}

\section{A monotone solution in pebble motion graphs}\label{sec:positive:assumption}
\vspace{-2pt}

\ifthenelse{\boolean{isfinalpaper}}{ %
To state our assumption for MSR instances
we define a monotone version of a Pebble Motion Graph (PMG) problem~\cite{DBLP:conf/focs/KornhauserMS84}. %
A PMG is a simple graph $\motiong = (V,E)$, also called a \emph{motion graph}, where $V$ consists of start vertices ($S$), target vertices ($T$), with $|S|=|T|$, and non-endpoint vertices ($Z$).
Each robot $r$ is represented by a pebble that is initially located at a distinct start vertex $u \in S$ and needs to be moved to its respective target $v \in T$.
Such a move is allowed if there is a path from $u$ to $v$ whose vertices (except $u$) do not contain a pebble.
Note that while the location of a pebble changes after each move, \motiong remains the same.
A solution to the PMG problem brings all the pebbles to their respective targets using one move at a time, with at most $|S|$ moves.
}{
    To state our assumption for MSR instances for uniform-shaped robots we study the underlying graph of Monotone MAPF instances, the discrete version of MSR.
    Recall that in Monotone MAPF the robots operate on a Pebble Motion Graph (PMG) $\motiong = (V,E)$, which we also call a \emph{motion graph} here,  where $V$ consists of start vertices ($S$), target vertices ($T$), with ${|S|=|T|}$, and non-endpoint vertices ($Z$).
    Each robot $r$ is represented by a pebble that is initially located at a distinct start vertex $u \in S$ and needs to be moved to its respective target $v \in T$.
    Such a move is allowed if there is a path from $u$ to $v$ whose vertices (except $u$) do not contain a pebble.
    A solution to Monotone MAPF brings all the pebble robots to their respective targets using one move at a time, with at most $|S|$ moves.
}
We now define our assumption that guarantees a solution to the PMG problem.

\begin{definition}  \label{def:sffbt}
    We say that the subset of vertices $V' \subseteq V$ in $G$ is \emph{\solvable{}}  (\emph{\shortsolvable{}} or \emph{spannable} for short) if there exists a spanning forest of $\motiong[V']$, the subgraph of $\motiong$ induced by $V'$, where each tree $\T$ in the forest satisfies the following:  $\T$ is full binary tree (i.e., every node has either 0 or 2 children) and every leaf of $\T$ is adjacent to some vertex $v \in Z$.
\end{definition}
\vspace{-6pt}
\begin{definition}  \label{def:sffbtPMG}
    We say that $\motiong{}$ is \emph{\solvable{}} (\emph{\shortsolvable{}} or \emph{spannable} for short)
    if (i) $S$ and $T$ are disjoint, (ii) the set $ Z $ of non-endpoint vertices is connected, and (iii) each of $ S $ and $ T $ is \solvable{}.
\end{definition}

\Cref{fig:sffbt-grid} and \Cref{fig:intro:tight}~(right) illustrate the definitions.

\tikzset{
    srcB/.style={
        fill=blue!50
    },
    srcA/.style={
      fill=red!65
    },
    trgA/.style={
      draw=red, line width=0.5mm
    },
    trgB/.style={
      draw=blue, line width=0.5mm
    },
    edge/.style={
      -{Latex[length=1.5mm, width=2mm]},
      line width=0.3mm,
      draw=black
    }
  }
\begin{figure}[t]
    \centering
    \begin{minipage}[b]{0.45\textwidth}
        \centering
        \includegraphics[width=1.02\textwidth]{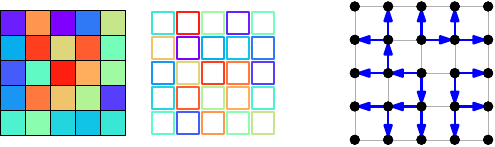}
        \caption{
        \MSRsq instance (left; same format as \Cref{fig:msr:intro}, cells of $Z$ not drawn) alongside (right) $G[S]$ and a spanning forest (blue arrows) showing that $S$ is \shortsolvable{}. As $G[T]$ has an analogous forest, the complete instance is spannable.
        }
        \label{fig:sffbt-grid}
    \end{minipage}
    \hfill
    \begin{minipage}[b]{0.5\textwidth}
        \centering
          \begin{minipage}{0.45\textwidth}
            \centering
            \begin{tikzpicture}[scale=0.6, font=\fontsize{8}{6.5}\selectfont, every node/.style={circle,draw},
          level 1/.style={sibling distance=2cm, level distance=0.7 cm},
          level 2/.style={sibling distance=1cm, level distance=1cm},
          inner sep=1pt] %
              \node[srcA](1) {1}
                child {node[srcB] (2) {2}
                  child {node[srcB] (4) {4}
                        child {node[srcA] (8) {8}}
                        child {node[srcB]  (9) {9}}
                        }
                  child {node[srcA] (5) {5}}
                }
                child {node[srcB] (3) {3}
                  child {node[srcB] (6)  {6}}
                  child {node[srcA] (7) {7}}
                };
            
              \path (2) edge[bend right=25,edge] (4);
              \path (4) edge[bend right=25,edge] (9);
              \path (3) edge[bend right=25,edge] (6);
            \end{tikzpicture}
          \end{minipage}
        \hfill  %
          \begin{minipage}{0.45\textwidth}
            \centering
            \begin{tikzpicture}[scale=0.6, font=\fontsize{8}{6.5}\selectfont, every node/.style={circle,draw}, level distance=1cm,
          level 1/.style={sibling distance=1.8cm, level distance=0.7 cm},
          level 2/.style={sibling distance=2cm, level distance=1 cm},
          level 3/.style={sibling distance=1cm,level distance=1 cm},
          inner sep=1pt]
              \node[trgB](4) {4}
                child {node[trgB] (3) {3}
                  child {node[trgB] (2) {2}
                        child {node[trgA] (7) {7}}
                        child {node[trgB]  (9) {9}}
                        }
                  child {node[trgA] (8) {8}
                        child {node[trgB] (6) {6}}
                        child {node[trgA]  (1) {1}}
                        }
                }
                child {node[trgA] (5) {5}} ;
            
              \path (5) edge[bend right=30,edge] (4);
              \path (8) edge[bend right=30,edge] (3);
              \path (1) edge[bend right=35,edge] (8);
              \path (7) edge[bend left=35,edge] (2);
            \end{tikzpicture}
          \end{minipage}
            \caption{ Illustration of \Cref{claim:spannable_feasible} for a forest with a tree for start positions (left) and a tree for targets (right), showing a partition of the robots into $B$ (blue) and $A$ (red).
            The $B$ robots move in the order 9,6,4,3,2 %
            to the target tree (via $Z$, not shown) using paths indicated by the arrows. %
            } \label{fig:spannable-motion}
    \end{minipage}
\end{figure}

\vspace{-4pt}
\subsection{Solving an instance with a spannable motion graph}\label{sec:positive:solve}
Solving the instances in our MSR hardness proofs boils down to choosing when the pivot robot $r^*$ moves with respect to other robots.
This choice is induced by a Pivot Scheduling problem where either $X_i$ or $Y_i$ contain multiple elements. %
We now show that spannable PMGs have a solution because they induce a Pivot Scheduling variant with singleton $X_i$'s and $Y_i$'s. We show that the latter always has a solution.

\begin{restatable}{lemma}{solvablepivot} \label{claim:pivot_scheduling_easy}
	Let $\P$ be a Pivot Scheduling instance where 
    $|X_i| = |Y_i| = 1$  for all $i$ and $|C_j| >1$ for all $j$.%
    \footnote{For this special case of Pivot Scheduling we write each before-constraint as  $B_i = \set{j_1, j_2}$ instead of $X_i = \set{j_1}$, $Y_i = \set{j_2}$.}
    Then $\mathcal{P} $ is feasible and can be solved in linear time.
\end{restatable}

\begin{maybeappendix}{solvablepivot}
\begin{proof} \vspace{-2pt}
    Given an instance $\P$ with the above properties we construct a SAT formula as follows.
    We create a variable $x_i$ for each before constraint $B_i$.
    We rename all the jobs so that the two jobs in each $B_i$ are $x_i$ and $\neg x_i$.
    With the jobs renamed, we create a clause for each after constraint $C_j$ which we take to be the conjunction of the literals represented by the jobs in $C_j$. %
    Due to the pairwise disjointness of the after constraints we obtain a SAT formula with non-singleton clauses where each variable appears at most twice (once negated and once unnegated).
    Such formulas always have a satisfying assignment, which can be found in linear time~\cite{DBLP:journals/dam/Tovey84}.
    Moreover, a satisfying assignment corresponds to a valid partition for $\P$.
    Specifically, if $x_i$ is assigned "true", then we include (the job of) $x_i$ in $A$ and $\neg x_i$ in $B$, otherwise we include  $\neg x_i$ in $A$ and $x_i$ in $B$.
    Clearly, $B$ hits all the $B_i$'s.
    Finally, $A$ hits all the $C_j$'s since each clause is satisfied and overall we get a valid partition for $\P$.
\end{proof}
\end{maybeappendix}

\ifthenelse{\boolean{ispaper}}{ %
The proof is via a reduction of $\P$ to a SAT formula that is always satisfiable; see \Cref{sec:missing}.
}{ %
}

\begin{theorem}\label{claim:spannable_feasible}
A PMG instance in which $\motiong$ is \shortsolvable{} has a solution.
If the corresponding spanning forest of \motiong is provided, then an ordering of the pebble robots in which they can be moved to their targets one by one can be found in $O(\numagents)$ time, for $\numagents$ pebbles.
\end{theorem} %
\begin{proof} \vspace{-2pt}
    Given such a motion graph \motiong{} along with the corresponding forest, we first show how to construct a Pivot Scheduling instance $\mathcal{P}$ that satisfies the requirements of \Cref{claim:pivot_scheduling_easy} (and is thus feasible). Then we apply \Cref{claim:pivot_scheduling_easy} to get a solution to $\mathcal{P}$, which we use to obtain a monotone motion plan.
	
    We define $\P$ as follows.
    For each robot we have a corresponding job in $\P$.
    For simplicity, we use the same notation to refer to a robot, its corresponding job, and nodes corresponding to the robot's two endpoints in $S$ and $T$.
    For each internal node $ u $ in the forest, let $ u_\ell $ and $ u_r $ be its left and right children.
    For each internal node $ u $ in the forest spanning $ S $, we have a before-constraint with $B_u = \set{u_\ell, u_r} $, 
    and for each internal node $ u'$ in the forest spanning $ T $, we have an after-constraint $ A_{u'} = \set{u'_\ell, u'_r} $.
    Note that there are no before-constraints (resp. after-constraints) on the roots of the trees of $S$ (resp. $T$) in the forest.
    It is straightforward to verify that $\P$ is solvable by \Cref{claim:pivot_scheduling_easy}.

    We now show how to use the solution to $\P$ to construct a monotone motion plan.
    Let $ (B, A) $ be a solution to $\P$.
    We first move the robots corresponding to $B$ one by one in increasing order of the height of their start nodes, i.e., in a bottom-up order.
    Since each parent node $u$ in $B$ imposes a before constraint on two of its children, at least one child appears before $u$ in the ordering. 
    Therefore, such an ordering allows us to move each robot $ u \in B $ out of the tree in which it starts to the connected set of vertices $Z$.
    Namely, $u$ can reach $Z$ by repeatedly moving from its present node to a child node corresponding to a robot of $B$ (which would be empty by the chosen order), until reaching a leaf, which is adjacent to a node of $Z$, by definition. See \Cref{fig:spannable-motion}.
    
    We now claim that there exists a robot-free path $\pi$ by which $u$ can continue its motion from $Z$ to its target position.
    The path $\pi$ exists since each target position $ v \in T $ corresponds either to a leaf in the forest (in which case it is directly reachable from $Z$) or to an after-constraint $A_v$.
    In the latter case, at least one of the children in $A_v$ does not belong to $B$, and, by induction on height, $v$ is again reachable from $Z$, regardless of which robots in $ B $ have already moved.
    More precisely, $\pi$ can be traced in reverse starting from the target of $u$ by repeatedly proceeding to a child node that is a target of a robot of $A$ (which is empty at this point since the robots of $A$ has not yet moved), until reaching a leaf.
    
    It remains to move all the robots corresponding to $ A $.
    We apply a symmetric argument to the one we have applied for the $ B $ robots, by using the after-constraints:
    Let us assume for a moment that all the robots are at their target positions.
    We can then move the robots of $A$ one by one to their start positions using the same argument as above, where we switch the roles of $A$ and $B$.
    Following the resulting order and corresponding paths in reverse will move the robots of $A$ to their targets in the original problem.
    The claim follows.
\end{proof}

\vspace{-6pt}
\subsection{Hardness of deciding whether a motion graph is spannable}
Deciding whether a given motion graph \motiong is spannable amounts to deciding whether a subset of its vertices $V' \subseteq V \setminus Z$ is spannable.
(We must decide this once for $V'=S$ and once for $V'=T$. The other requirements in \Cref{def:sffbtPMG} are trivial to verify, so we assume they hold.) %
To study the latter, it is convenient to consider the following problem.

\noindent
\textbf{Leaf-Constrained Spannability (LCS) problem.} Given $(H, L)$, where $H = (V_H, E_H)$ is a connected simple graph and $L \subseteq V_H$ is a subset of its vertices, determine whether $H$ has a spanning forest consisting of full binary trees such that every leaf in the spanning forest is in $L$.
We refer to a "yes" instance of LCS as \emph{spannable}.

Determining whether \motiong is spannable equates to solving an induced set of LCS instances as formalized by \Cref{lem:inducedLCS}, which follows directly from our definitions.

\begin{lemma} \label{lem:inducedLCS}
Let $H_1, \ldots, H_q$ be the connected components of $\motiong[V']$.
For each $H_i$, let $L_i$ denote the subset of vertices of $H_i$ that are adjacent to some vertex $v \in Z$ in \motiong.
Then $V'$ is spannable if and only if for each $i$ the LCS instance $(H_i,L_i)$ is spannable.
\end{lemma}

We call a vertex $v$ a \emph{boundary} vertex if $v \in L$; otherwise, it is an \emph{inner} vertex. 
The \emph{depth} of a vertex $v$ is the shortest path distance from $v$ to the nearest boundary vertex.
The \emph{depth} of an LCS instance is the maximum depth of a vertex in $V_H$.
\ifthenelse{\boolean{ispaper}}{ %
We prove that for general graphs LCS is NP-complete using a reduction from Hamiltonian Path.
Fortunately, this hardness result does not apply to graphs induced by geometric instances, as we demonstrate in the sequel using the concept of depth.
}{ %
}

\begin{restatable}
{theorem} {LCSHardness} 
LCS is NP-complete even for a depth of 1.
\end{restatable}
\begin{maybeappendix}{LCSHardness}
\begin{proof} \vspace{-2pt}
We perform a reduction from Hamiltonian Path.
We turn a given graph $G$ into a graph $H$ for the LCS instance as follows:
Add a new vertex $w$ adjacent to all vertices in $G$.
Then, for each vertex $ v \in G$, $v \neq w$, add an adjacent leaf vertex (i.e., having $v$ as its only neighbor).
We set $L$ to be all the newly added vertices.
It is straightforward to identify a Hamiltonian path in $G$ with a spanning tree for $H$ as required for LCS.
\end{proof}
\end{maybeappendix}

\ifthenelse{\boolean{ispaper}}{ %
}{ %
Fortunately, this hardness result does not apply to graphs induced by geometric instances, as we demonstrate next using the concept of depth. 
}

\vspace{-5pt}
\section{Application in geometric settings} \label{sec:positive:spannability}
\vspace{-4pt}
We now apply our positive result for PMGs to geometric and non-discrete instances.
We begin with MSR for unit-disk robots operating in the plane.
Let \msrdisks be such an instance with $S$ and $T$ denoting the start and target positions, respectively, where a \emph{position} is specified by the location of a robot's center.
Our approach consists of constructing a motion graph $\motiongdisks \coloneqq \motiong(\msrdisks)$ such that a solution for \motiongdisks can be translated to a solution for \msrdisks.
We take the vertices of \motiongdisks to be $V = S \cup T \cup \set{z}$, i.e., we consider each start/target position as a vertex and use an additional position $z$ as an empty vertex.
Hereafter we will typically not distinguish between a position and its corresponding vertex in \motiongdisks.
Our goal is to associate each edge $e=(u,v)$ in \motiongdisks with a corresponding path $\pi(e)$ from $u$ to $v$ in \msrdisks such that a motion of a pebble robot along $e$ always corresponds to a collision-free motion along $\pi(e)$.
Therefore, we add edge $e$ to \motiongdisks if and only if path $\pi(e)$ exists in \msrdisks such that a robot moving along $\pi(e)$ will not collide with a robot located at any position $w \in (S \cup T) \setminus \{u,v\}$.
We will show how to compute efficiently such edges, which are determined uniquely based on $V$.
Then, we will show that we can efficiently determine whether \motiongdisks is spannable by exploiting the properties of its induced LCS instances.

Since our main goal is to show that spannability can be efficiently decided for PMGs based on non-discrete inputs, we do not tighten running times and assume no obstacles, which can be handled efficiently but are omitted for simplicity.

\vspace{-4pt}
\subsection{Preliminaries and background}\label{sec:prelim}
\vspace{-2pt}
For a point $p \in \dR^2$, let $\D_r(p)$ denote the open disk of radius $r$ centered at point $p$.
Let $\E = \bigcup_{x \in S \cup T} \D_1(x)$ be the union of all unit disks positioned at endpoints.
Let $\Z = \{x \in \dR^2 : \D_1(x) \cap \E = \emptyset \}$ %
denote the \emph{endpoint-free space}, i.e., all positions where a unit-disk robot is guaranteed not to intersect another robot located at an endpoint. %
To ensure conditions (i) and (ii) of \Cref{def:sffbtPMG} we require the following in \msrdisks:
(i) unit disks placed at $S \cup T$ are pairwise interior-disjoint and (ii) $\Z$ is a single connected component.
We can easily test these conditions as part of the construction of \motiongdisks (to be described) and report if they do not hold.

\noindent
\textbf{Arrangements.} An arrangement of circles is a subdivision of the plane into vertices (intersection points of the circles), edges (maximal arcs of a circle not intersected by any other circle), and faces (maximal portions of the plane that are not intersected by any circle) induced by the given circles.
For ease of exposition, we assume general position, i.e., no more than two circles intersect at any given point and there are no two tangent circles.

\noindent
\textbf{Courcelle's theorem.}
To show that LCS can be efficiently decided for geometric inputs we require a basic understanding of treewidth and Courcelle's theorem~\cite{DBLP:journals/iandc/Courcelle90}.
In brief, \emph{treewidth} measures how "tree-like" a graph is using a natural number.
Courcelle's theorem states that any graph property that can be expressed in Monadic Second-Order logic (MSO) is decidable in linear time on graphs with a fixed treewidth.
See~\cite{cygan2015parameterized} for a more comprehensive background.

\vspace{-6pt}
\subsection{Constructing the motion graph}\label{sec:positive:constructingG}
\begin{figure}[t]
    \centering
    \begin{minipage}[b]{0.495\textwidth}
        \centering
        \includegraphics[width=0.94\textwidth]{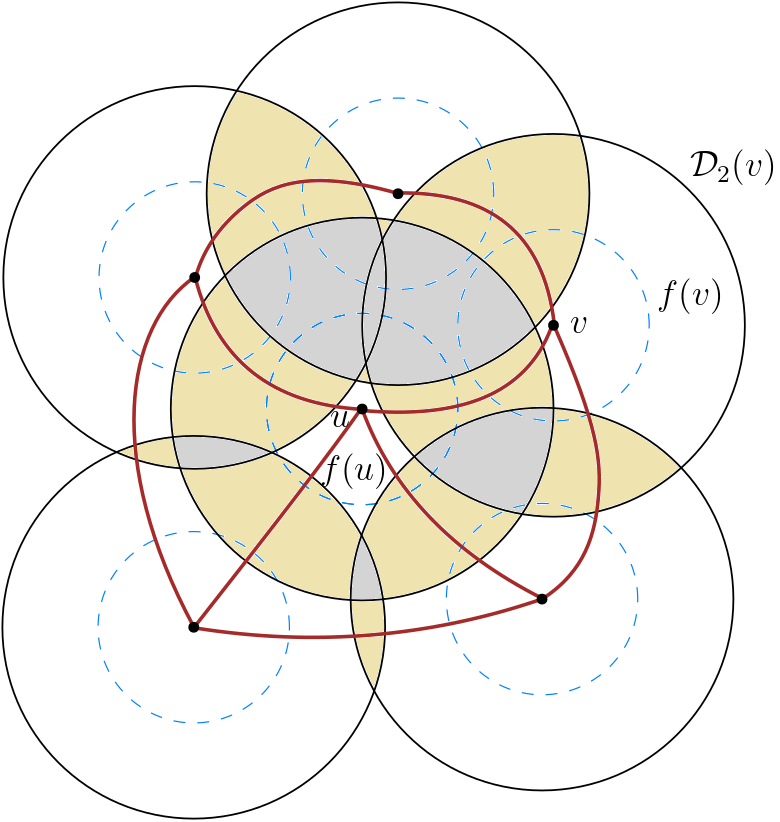}
        \caption{
        Partial view of $\A$ showing robots at their endpoints (blue dashed circles), corresponding circles of $\C$ (large black circles), and edges of \motiongdisks {} (thick brown curves) between shown endpoints. Faces of $\A$ contained in 2 (resp. more than 2) disks are shaded in beige (resp. gray).
        }
        \label{fig:arr}
    \end{minipage}
    \hfill
    \begin{minipage}[b]{0.46\textwidth}
        \centering
        \includegraphics[width=1\textwidth,angle=90,origin=c]{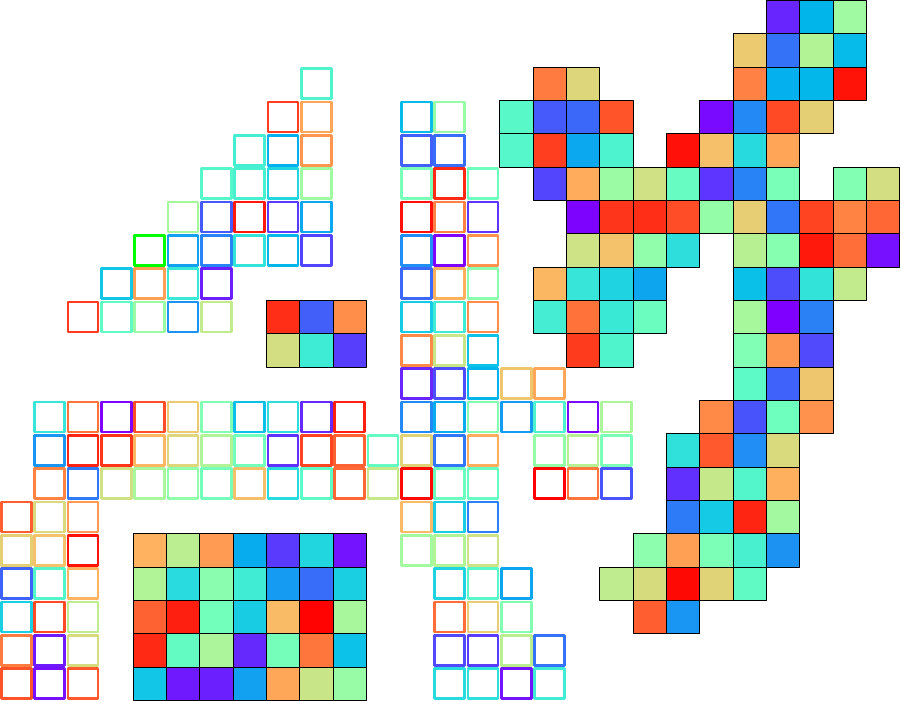}
        \caption{\MSRsq instance, which is \shortsolvable{}, composed of $5$ polyominoes satisfying conditions (i)-(iii) in \Cref{sec:positive:polyomino} except for the $5\times7$ polyomino (right), which violates (ii), but is nevertheless spannable. 
        Filled (resp. unfilled) cells are start (resp. target) positions. The rest of the grid cells (not drawn) are empty.
        }
        \label{fig:examplespoly}
    \end{minipage}
\end{figure}

\vspace{-4pt}
To construct \motiongdisks we subdivide the plane into maximal path-connected regions, where a robot positioned anywhere in a region intersects the same subset of endpoints. %
This is done by computing the arrangement $\A \coloneqq \A(\C)$ where $\C$ is the set of radius 2 circles centered at $S \cup T$; see \Cref{fig:arr}.
Notice that the circles of $\C$ that contain a point $p$ correspond to the endpoints that a robot centered at $p$ intersects with.
Also, since $\Z$ is a single connected component, $\Z$ corresponds to the single unbounded face of $\A$.
Let us fix $z$ as an arbitrary position in $\Z$.

We now construct \motiongdisks. Since we fix the vertices to be $V = S \cupdot T \cupdot \set{z}$,
the main task is determining which edges to add.
For $v \in V$, let $f(v)$ denote the face of $\A$ containing $v$. %
We must ensure that the motion of a pebble across an edge $(u,v)$ in \motiongdisks corresponds to a collision-free motion of a unit disc between the two positions in \msrdisks.
Hence, for $u,v \in S\cup T$, we will add $(u,v)$ to \motiongdisks if and only if a path $\pi$ exists from $u$ to $v$ that does not intersect any disk $\D_2(w)$ for $w \in (S \cup T) \setminus \{u,v\}$, i.e., a robot can move along $\pi$ without colliding with other robots located at positions $V \setminus \{u,v\}$.
It suffices to consider the case where $\pi \subset \D_2(u) \cup \D_2(v)$.%
\footnote{The only other option is when $\pi$ passes through $\Z$. In this case the edges $(u,z)$ and $(v,z)$ would be present in \motiongdisks, which would make $u$ and $v$ boundary vertices in any LCS instance induced by \motiongdisks.
Since any spannable LCS instance remains spannable if we remove edges between two boundary vertices, we can safely ignore $(u,v)$.} %
In this case, $\pi$ passes through exactly 3 faces,
which are $f(u)$, $f(v)$, and a face contained in
$\D_2(u) \cap \D_2(v)$ but not in any other circle.
The other type of edge, $(u,z)$, is added to \motiongdisks if and only if the faces $f(u)$ and $f(z)$ are adjacent, i.e., they share a common boundary.

Following our observations, a straightforward traversal of $\A$ obtains the edges of \motiongdisks{}, for which we can also obtain the corresponding paths in \msrdisks.
We can construct $\A$ using a standard sweep-line algorithm in $O(\numagents \log \numagents)$ time, and traverse it in the same time~\cite{CGALarrangements}. %
\ifthenelse{\boolean{ispaper}}{ %
}{%
(We provide low-level details and a running time analysis in the proof of \Cref{thm:msrdisks} below.)%
}
Notice that \motiongdisks{} is planar and has a complexity of $O(\numagents)$ since the dual graph\footnote{In the dual graph of $\A$ each node represents a face, and an edge connects two nodes if their corresponding faces are adjacent in $\A$.}
of $\A$ is planar and \motiongdisks{} can be obtained from the dual graph by deleting vertices and contracting edges.

\vspace{-5pt}
\subsection{Efficiently deciding whether the motion graph is spannable}\label{sec:positive:decision}
Let $(\hat{H}, L)$ be an LCS instance induced by \motiongdisks (cf. \Cref{lem:inducedLCS}).
We now show that we can solve $(\hat{H}, L)$ in linear time.

\begin{lemma} \label{lem:bounded-depth}
If $(\hat{H}, L)$ is spannable then its depth is at most 9.
\end{lemma}
\begin{proof} \vspace{-2pt}
Let $d$ be the depth of $(\hat{H}, L)$ and let $v \in V_{\hat{H}}$ be a vertex with such a depth.
Notice that the forest spanning $\hat{H}$ contains a subtree $\T$ rooted at $v$, which is a perfect binary tree of height $d$, i.e., $\T$ has $2^{d+1}-1$ nodes.
This is true since the depth of a node in the forest differs from the depth of its parent by at most 1 and each tree in the forest is full.
Let us now bound the area available for the start/target positions of the unit-disk robots corresponding to the nodes of $\T$.
For any $(u,v) \in E_{\hat{H}}$, which is also an edge of \motiongdisks{}, the distance between positions $u$ and $v$ is at most $4$ as $\D_2(u)$ and $\D_2(v)$ must intersect for $(u,v)$ to exist. %
Consequently, any robot in $\T$ is positioned at most $4d$ units from $v$.
That is, all robots in $\T$ must be located inside $\D_{4d}(v)$. %
The total area of ${2^{d+1}-1}$ unit disks exceeds the area of $\D_{4d}(v)$ for $d > 9$, which concludes the argument.
\end{proof}

By \Cref{lem:bounded-depth} we easily rule out $(\hat{H}, L)$ as not spannable if its depth is greater than 9 using a breadth-first search from $L$.
\ifthenelse{\boolean{ispaper}}{ %
Otherwise, $\hat{H}$ has a constant depth, which, together with its planarity, means that $\hat{H}$ has a constant treewidth~\cite{DBLP:journals/tcs/Bodlaender98}.
We therefore apply Courcelle's theorem in the latter case to show that we can efficiently decide spannability. %
To do so, it suffices to express spannability as an MSO formula that is true if and only if $(\hat{H}, L)$ is spannable.
We may construct such a formula having a constant size using standard predicates.
}{ %
Otherwise, $\hat{H}$ has a constant depth and is therefore efficiently solvable by the following lemma.
}

\begin{restatable}
{lemma}{decidingLCSgeo}  \label{lem:decidingLCSgeo}
Let $(H, L)$ be an LCS instance where $H$ is planar and has a constant depth. Then we can solve $(H, L)$ in linear time.
\end{restatable}

\begin{maybeappendix}{decidingLCSgeo}
\begin{proof} \vspace{-2pt}
Since $H$ is planar and has a constant depth $c$, it is $(c+1)$-outerplanar.%
\footnote{A graph $G$ is $k$-outerplanar if it has a planar embedding such that every vertex in $G$ can be reached from the unbounded face by traversing an alternating sequence of at most $k$ faces and $k$ vertices, where each consecutive face and vertex are incident to each other.}
Thus, $H$ has a constant treewidth since any $c'$-outerplanar graph has a constant treewidth when $c'$ is a fixed constant~\cite{DBLP:journals/tcs/Bodlaender98}.
We therefore apply Courcelle's theorem, for which it suffices to express the spannability property in a Monadic Second-Order logic (MSO) formula. %
The formula can be expressed as follows:
\begin{align*}
\exists E' \subseteq E_{\lcsdisks{}}, R \subseteq V_{\lcsdisks{}} \, \bigl( \text{Forest}(E', R) \land 
\forall v \in V, \exists r \in R \, \bigl( &\text{Reachable}(v, r, E') \land \quad  \\
&\text{UniqueParent}(v, E') \land \quad \\
&\left( v \in L \lor \text{TwoChildren}(v, E') \right) \bigr) \bigr)
\end{align*}

where the auxiliary formulas are defined as:
\begin{itemize}
    \item \(\text{Forest}(E', R)\): The edges in \( E' \) form a forest with roots in set \( R \).
    \item \(\text{Reachable}(v, r, E')\): Vertex \( v \) is reachable from vertex \( r \) using only edges in \( E' \).
    \item \(\text{UniqueParent}(v, E')\): Vertex \( v \) has exactly one parent in the forest formed by \( E' \).
    \item \(\text{TwoChildren}(v, E')\): Vertex \( v \) has exactly two children in the forest formed by \( E' \).
\end{itemize}

The auxiliary formulas can be expressed using standard constructs~\cite{cygan2015parameterized}.
\end{proof}
\end{maybeappendix}

Putting everything together, we obtain the following.
\begin{restatable}
{theorem}{unitdisks} 
\label{thm:msrdisks}
Let $M$ be an MSR instance for $\numagents$ unit disks in the plane.
We can construct the motion graph \motiongdisks{} for $M$, decide whether \motiongdisks{} is spannable, and if so output a solution for $M$, in $O(\numagents^2)$ time.
\end{restatable}
\begin{maybeappendix}{unitdisks}
\begin{proof}
We first construct the arrangement $\A$ in $O(\numagents \log \numagents)$ time using a standard sweep-line algorithm.
The running time follows from the fact that the complexity of $\A$ is $O(\numagents)$. This is true since unit disks placed at $S \cup T$ are pairwise interior-disjoint, and so each circle in $\C$ intersects a constant number of circles in $\C$.
We then traverse $\A$ as follows to determine the edges of \motiongdisks.
We examine all faces contained in exactly two disks. For such a face $f$ contained in $\D_2(u) \cap \D_2(v)$, we examine whether $f$ is adjacent to $f(u)$ and $f(v)$, and add the edge $(u,v)$ if so.
Note that the disks in which the currently visited face is contained are readily maintained by annotating each edge (i.e., circular arc) of $\A$ with its inducing circle of $\C$.
The traversal infers the adjacencies of all faces, thereby also determining the edges of \motiongdisks{} incident to $z$.
After obtaining \motiongdisks, we use \Cref{lem:bounded-depth} and \Cref{lem:decidingLCSgeo} to determine whether it is \shortsolvable.
If that is the case, we apply \Cref{claim:spannable_feasible} to get a monotone solution to \motiongdisks.

\ifthenelse{\boolean{ispaper}}{ %
}{ %
\begin{figure}[t]
  \centering
  \includegraphics[width=0.5\textwidth]{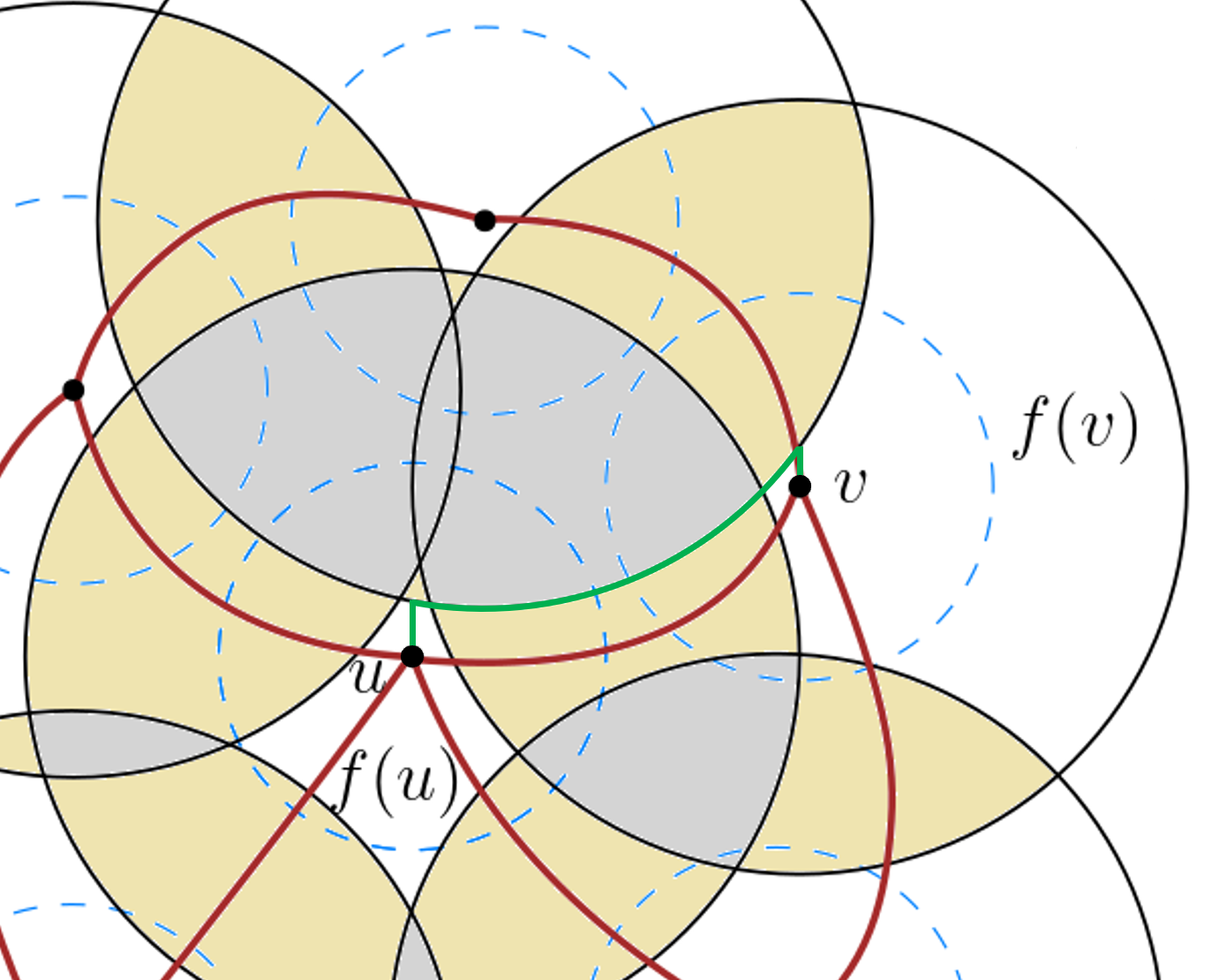}
 \caption{ The path $\pi(e)$ (green) corresponding to the edge $e=(u,v)$ for the example in \Cref{fig:arr}.}
\label{fig:arr_path}
\end{figure}
}

By the construction of \motiongdisks, each move of a pebble to its target via a path $\gamma$ in \motiongdisks corresponds to a collision-free path $\pi$ in \msrdisks.
We now describe how to find such a path $\pi$ consisting of $O(\numagents)$ line segments and circular arcs.
For each edge $e=(u,v)$ along $\gamma$ we associate a path $\pi(e)$ in \msrdisks as follows.
Let us first assume $z \notin \set{u,v}$, in which case $\pi(e)$ traverses the faces $f(u),f',f(v)$, where $f'$ is a face contained in $\D_2(u) \cap \D_2(v)$.
To obtain the first segment of $\pi(e)$ we follow a vertical ray upwards from $u$ until it hits the boundary of the face $f(u)$.
We then follow the boundary of $f(u)$ to the boundary of $f'$ (these boundaries must partially overlap since the faces are adjacent) and then continue similarly to the boundary of $f(v)$.
From the latter boundary, we obtain the last segment of $\pi(e)$ using a ray similarly to its first segment.
\ifthenelse{\boolean{ispaper}}{%
}{%
See \Cref{fig:arr_path}.%
}
In the other case, we have $v=z$.
The first segment of $\pi(e)$ connects to the boundary of $f(u)$ as in the first case.
We then proceed along the boundary to a vertex $p$ of $\A$ that is incident to the unbounded face.
The path $\pi(e)$ then follows the line segment $\overline{pz}$ towards $z$, with detours as follows.
For each maximal portion $\overline{qq'}$ of $\overline{pz}$ that penetrates bounded faces, we traverse the boundary of the unbounded face from $q$ to $q'$ and continue along $\overline{pz}$.
We can compute $\pi$ in $O(\numagents)$ time given $\A$ and output all the paths in $O(\numagents^2)$.
\end{proof}
 \end{maybeappendix}

\subsection{A restricted case that always has a monotone solution}\label{sec:positive:polyomino}
We now identify a restricted variant of \MSRsq
that always has a solution.
Let \motiong be a PMG of an \MSRsq instance in which $Z$ is connected.
We require the following for each connected component $P$ of $S$ or $T$ in \motiong, which is commonly called a \emph{polyomino}:\footnote{We do not distinguish between a vertex of \motiong and its corresponding grid square.} (i) $P$ is simple, i.e., it has no holes, (ii) each square in $P$ is either a boundary square  (i.e., a square with an edge on the perimeter) or is adjacent to a boundary square, and (iii) each boundary square of $P$ is adjacent to a square of $Z$.
We say that such a PMG \motiong \emph{\thinpolys}; see \Cref{fig:examplespoly} for an example.

\begin{restatable}
{theorem}{thinpoly} \label{thm:thinpolys}
 A PMG \motiong that \thinpolys{} is spannable and thus solvable. %
\end{restatable}

\ifthenelse{\boolean{ispaper}}{ %
We prove \Cref{thm:thinpolys} through a careful analysis showing that each polyomino $P$ contains a \emph{good square}, which is an inner square $s$ that has two adjacent boundary squares in $P$ whose sole adjacent inner square is $s$; see \Cref{fig:good-square}(a) in \Cref{sec:missing}. 
We then recursively get an appropriate spanning tree for $P$ by removing such a square.
}{}

\begin{maybeappendix}{thinpolydetails}
Towards proving \Cref{thm:thinpolys} first notice that for such a graph \motiong the following holds for each induced LCS instance $(H, L)$ of \motiong (cf. \Cref{lem:inducedLCS}) by definition: $H$ is a simple polyomino $P$ and has a depth of at most 1, where $L$ consists exactly of the boundary squares of $P$.
As it is straightforward to identify $(H, L)$ with $P$, in the sequel we treat $P$ as an LCS instance and do not distinguish between the two.
We now establish two lemmas from which \Cref{thm:thinpolys} follows.

We say that an inner square $s$ in $P$ is \emph{good} if there are two boundary squares in $P$ whose sole neighboring inner square is $s$; see \Cref{fig:good-square}(a).

\begin{figure}[t]
    \centering
    \includegraphics[width=0.6\textwidth]{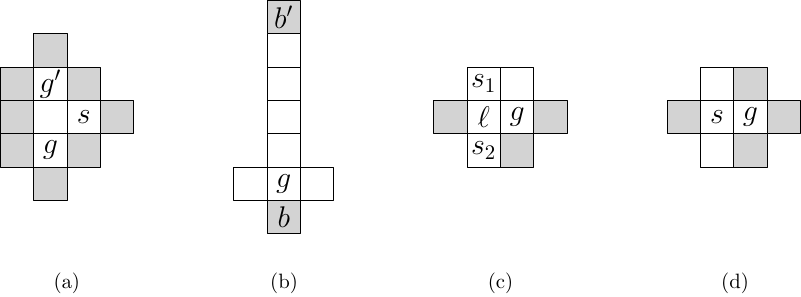}
    \caption{(a) A polyomino in which the inner squares $g$ and $g'$ are good but $s$ is not. Boundary squares are shaded.
    (b)-(d) The cases arising in the proof of \Cref{lem:good-square}.}
    \label{fig:good-square}
\end{figure}

\begin{restatable}{lemma}{goodsquare} \label{lem:good-square}
A simple polyomino $P$ with a depth of 1 has a good square.
\end{restatable}

\begin{proof} \vspace{-2pt}
Let $Q$ be a subpolyomino restricted to the inner squares of $P$.
We fix an arbitrary anchor square $a \in Q$ and let $g \in Q$ denote the farthest square from $a$ in $P$ among all squares in $Q$.
Let us assume that $Q$ is connected, as the remaining cases are similar. %

First we show that $g$ cannot be adjacent to exactly one boundary square. Suppose that this is the case. Without loss of generality the single boundary square $b$ adjacent to $g$ is below it; see \Cref{fig:good-square}(b).
Let $b'$ be the first boundary square intersected by a ray emanating upwards from $g$ and let $C$ denote the column of inner squares between $b$ and $b'$.
If $a \in C$ then clearly one of the horizontal neighbors of $g$ is farther from $a$ than $g$.
Otherwise, $C$ splits $Q$ into multiple components, one of which contains $a$.
Without loss of generality, $a$ resides in a component connected to $C$ from the left. Then, the right neighbor of $g$ is farther from $a$ than $g$.
Thus, all cases contradict the choice of $g$.
We refer to the argument above as a \emph{cut argument}, which we will use below without full reiteration.

Next, we examine the case where $g$ is adjacent to exactly two boundary squares. If these two boundary squares lie on the same row or column, then we get a similar contradiction to the choice of $g$ as in the previous case.
Hence, we assume without loss of generality that $g$ has a neighboring boundary square below it and another one to its right; see \Cref{fig:good-square}(c).
Let $\ell$ be the left neighbor of $g$.
Then, $\ell$ has an inner square $s_1$ as a neighbor above it, as shown, since otherwise $g$ cuts $Q$ in contradiction to the choice of $g$ by the cut argument.
Now let us assume that $g$ is not a good square, i.e., there exists an inner square $s_2$ adjacent to one of the boundary-square neighbors of $g$.
Without loss of generality, $s_2$ is located below $\ell$, as shown. %
Since $\ell$ has a depth of $1$, its left neighbor must be a boundary square.
Consequently, $\ell$ splits $Q$ into two components. We cannot have $a$ be in the same component as $s_2$, as that would contradict the choice of $g$ as the farthest square from $a$.
To avoid the same contradiction, the component containing $s_2$ must be the singleton $\set{s_2}$.
Therefore, $s_2$ is adjacent to 3 boundary squares and must be a good square.

In the remaining case, $g$ is adjacent to exactly 3 boundary squares. Without loss of generality, the square to the left of $g$ is an inner square $s$; see \Cref{fig:good-square}(d). If $g$ is not a good square, then the two vertical neighbors of $s$ must be inner squares.
Since $s$ has a depth of $1$, its left neighbor must be a boundary square.
If $a = s$ then $Q$ consists exactly of the inner squares identified so far and thus both vertical neighbors of $s$ are good squares.
Otherwise, $s$ cuts $Q$ similarly to $\ell$ in the previous case, and we may use a similar argument to conclude that one of the vertical neighbors of $s$ is good.
\end{proof}

\ifthenelse{\boolean{ispaper}}{ %
}{ %
}

\begin{restatable}{lemma}{polyspannable} \label{thm:poly-spannable}
An LCS instance induced by a simple polyomino $P$ with a depth of at most 1 is spannable.
\end{restatable}

\begin{proof} \vspace{-2pt}
We prove this by induction on the number of inner squares. If there are no inner squares, then $P$ is spannable since each square can be considered a singleton tree.
Otherwise, by \Cref{lem:good-square}, there is a good square $g$ with two accompanying boundary squares $b_1$ and $b_2$.
Let $P'$ be the polyomino obtained by removing $b_1$ and $b_2$ from $P$.
Since $g$ is the only inner square adjacent to each of $b_1$ and $b_2$, every inner square in $P$ remains an inner square in $P'$, except for $g$, which becomes a boundary square.
Hence, $P'$ has one fewer inner square than $P$ and remains a simple polyomino with a depth of at most 1, making it spannable by induction.
Let $\T$ be the tree of the spanning forest of $P'$ that contains $g$ as a leaf.
We augment $\T$ by making $b_1$ and $b_2$ the two children of $g$, thus obtaining a spanning forest for $P$.
\end{proof}
\end{maybeappendix}

\section{Conclusion}
Through a detailed investigation of MSR we identify a new structural assumption, \shortsolvable, guaranteeing a solution, which we show can be efficiently decided on geometric MSR inputs.
SFFBT relaxes existing separation assumptions guaranteeing a solution in complete efficient MRMP algorithms.
We thereby echo calls for establishing such milder conditions~\cite{DBLP:journals/tase/AdlerBHS15, DBLP:conf/compgeom/BanyassadyBBBFH22} and give insights into the challenging complexity landscape of MRMP in denser scenarios.
Roughly speaking, we allow a start/target position to be closely surrounded by other such positions, provided that it is not too deeply packed among them.
\Cref{fig:examplespoly} illustrates the variety of tight configurations that can arise in spannable \MSRsq instances. In particular, a robot may be tightly surrounded by two "layers" of other robots as in the $7 \times 5$ polyomino shown.
SFFBT also generalizes the notion of well-formed environments~\cite{DBLP:journals/tase/Cap0KS15}, where each endpoint can be seen as a singleton tree in \Cref{def:sffbtPMG}.

We remark that a solution to MSR can apply to settings where parallel motion is allowed (e.g., MRMP).
One could reduce the reconfiguration time of a monotone solution by repeatedly replanning the trajectory of small subsets of the robots while treating the rest as moving obstacles, akin to prioritized planning~\cite{DBLP:journals/tase/Cap0KS15}.

\textbf{Additional results.}
Our negative and positive results apply to two related problems for which MSR is a special case.
On the negative side, we establish the NP-hardness of optimally decoupling an MRMP instance into sequential plans~\cite{DBLP:conf/rss/BergSLM09} (we recall its definition in \Cref{sec:opt-decoupling-def})
and strengthen the NP-hardness of Reconfiguration with Minimum Number of Moves~\cite{DBLP:journals/comgeo/DumitrescuJ13}.
Specifically, while the proof of~\cite{DBLP:journals/comgeo/DumitrescuJ13} relies heavily on obstacles, we present hardness without obstacles.
Also, it may be verified that all of our reductions have a linear blow-up.
Consequently, they establish a $2^{\Omega(\numagents)}$ running time lower bound, for \numagents objects, conditioned on the
Exponential Time Hypothesis (ETH)~\cite{DBLP:journals/jcss/ImpagliazzoP01}, which is a first for the latter problem (the blow-up in~\cite{DBLP:journals/comgeo/DumitrescuJ13} is at least quadratic).
In this regard, our running time lower bound matches the upper bound, up to a polynomial factor, of the dynamic programming-based MSR algorithm in~\cite{wang2021uniform}, establishing its optimality.
On the positive side, our results also apply to the two related problems since a monotone solution is necessarily optimal for both problems, as is readily verifiable.

\textbf{Future work and open problems.}
First, our result for efficiently deciding whether unit disc MSR instances are spannable can be similarly applied to other uniform shapes (e.g., unit squares).
We believe that \shortsolvable can be generalized in a few other ways.
One could allow multiple components of $Z$, whereby a robot would traverse many trees to reach its target.
Another generalization is trees containing both start and target positions.

A more direct efficient LCS algorithm for geometric inputs would be more practical than applying Courcelle's theorem.
One alternative is a standard dynamic programming approach for bounded treewidth graphs~\cite{cygan2015parameterized}.
Better yet, can we characterize spannable graphs such as, e.g., those arising in \MSRsq?
A necessary condition is that there are fewer inner than boundary squares, however it is not sufficient.%

For the cases where we show hardness for \MSRnear, what is the complexity for \MSRsq?
Specifically for the latter, consider $n^2$ robots where each of the start and target configurations lies on a $n \times n$ portion of the 2D grid, and the configurations are separated by an empty column (generalizing \Cref{fig:sffbt-grid}).
Are such instances NP-hard to solve? Do negative instances even exist?

\bibliographystyle{plainurl}%
\bibliography{references}

\appendix

\section{Missing proofs} \label{sec:missing}

\pivot*
\begin{proof}
We present a reduction from 3SAT, the problem of deciding satisfiability of a formula in conjunctive normal form with three literals in each clause.
Given a 3SAT formula $\phi$, we define a corresponding instance of Pivot Scheduling $\P \coloneqq \P(\phi) = (V, \C)$, which has a valid partition if and only if $\phi$ is satisfiable.

For each appearance of a literal in $\phi$ we have a corresponding job and each $X_i, Y_i$ pair in $\P$ contains jobs corresponding to one variable.
That is, for each variable $x_i$, we add to $X_i$ (resp. $Y_i$) a job for each appearances of $x_i$ (resp. $\overline{x_i}$) in $\phi$.
As for after constraints, each clause of $\phi$ defines a constraint $C_j$, which contains the jobs corresponding to the particular appearances of literals in the clause.
It is easily seen that any pair of the sets $X_i$'s and $Y_i$'s is disjoint and also that the $C_j$'s are disjoint.

To complete the reduction, we define the correspondence between a satisfying assignment $\A$ for $\phi$ and a valid partition of $V$:
A variable $x_i$ is assigned "true" if and only if $X_i \subset \aft$.
In other words, $\bef$ and $\aft$ correspond to literals that evaluate to "false" and "true" according to $\A$, respectively.
It is straightforward to verify that the resulting partition satisfies the constraints if and only if $\A$ satisfies $\phi$.

To show that the problem remains hard when $|X_i| =3, |Y_i|=1$ for all $i$, we perform the same reduction from a special version of 3SAT, in which each variable appears exactly three times unnegated and once negated, which remains NP-hard~\cite{DBLP:journals/dam/DarmannD21}. %
\end{proof}

\putmaybeappendix{solvablepivot}

\putmaybeappendix{LCSHardness}

\putmaybeappendix{decidingLCSgeo}

\putmaybeappendix{unitdisks}

\putmaybeappendix{thinpolydetails}

\section{Optimal decoupling into sequential plans}\label{sec:opt-decoupling-def}
We recall the definition of the problem of optimal decoupling into sequential plans following~\cite{DBLP:conf/rss/BergSLM09}.
We are given an MRMP instance $\I$ consisting of a set $R$ of robots, each with its own start and target positions.
A \textit{decoupled execution sequence} is an ordered partition $S=(R_1, \ldots, R_k)$ of $R$, i.e., $R_i \cap R_j = \emptyset$ for $i \neq j$ and $\bigcup_i R_i = R$.
Such a sequence is a \textit{solution sequence} for $\I$ if there is a motion plan bringing the robots to their respective targets (without causing collisions) that can be divided into $k$ time intervals $\tau_1, \ldots, \tau_k$ where at each interval $\tau_i$ only the robots in $R_i$ move.
The dimension of a solution sequence $S$, denoted by $\dim(S)$, is the size of the largest set of robots in $S$, i.e., $\dim(S) \coloneqq \max_{R_i \in S} |R_i|$.
A solution sequence $S$ is \emph{optimal} if it has the smallest dimension over all possible solution sequences for $\I$.
The problem asks to find an optimal solution sequence.

\end{document}